\pdfoutput=1
\documentclass[11pt]{article}

\usepackage[final]{acl}

\usepackage{times}
\usepackage{latexsym}

\usepackage[T1]{fontenc}
\usepackage{CJKutf8}
\usepackage[utf8]{inputenc}
\usepackage{amsmath, amssymb}

\usepackage{booktabs}

\usepackage{caption}
\usepackage{subcaption}

\usepackage{microtype}

\usepackage{inconsolata}

\usepackage{graphicx}

\usepackage[normalem]{ulem}
\useunder{\uline}{\ul}{}
\usepackage{multirow}
\usepackage{changepage}
\usepackage{placeins}
\usepackage{url}

\title{Are Generative Models Underconfident? Better Quality Estimation with Boosted Model Probability}

\author{Tu Anh Dinh \and Jan Niehues \\
        Karlsruhe Institute of Technology \\ Karlsruhe, Germany \\ \texttt{\{firstname\}.\{lastname\}@kit.edu}}

\begin{document}
\maketitle
\begin{abstract}
Quality Estimation (QE) is estimating the quality of the model output during inference when the ground truth is not available. Deriving output quality from the models' output probability is the most trivial and low-effort way. However, we show that the output probability of text-generation models can appear \textbf{underconfident}. At each output step, there can be multiple correct options, making the probability distribution spread out more. Thus, lower probability does not necessarily mean lower output quality. Due to this observation, we propose a QE approach called \textbf{\textsc{BoostedProb}}\footnote{Implementation available at \url{https://github.com/TuAnh23/boostedprob}.}, which boosts the model's confidence in cases where there are multiple viable output options. With no increase in complexity, \textsc{BoostedProb} is notably better than raw model probability in different settings, achieving on average +0.194 improvement in Pearson correlation to ground-truth quality. It also comes close to or outperforms more costly approaches like supervised or ensemble-based QE in certain settings.

% when estimating the quality of different models (Whisper, Llama, etc.) on different tasks (translation, summarization, etc.). Compared to raw model probability, \textsc{BoostedProb} achieves on average +0.194 improvement in Pearson correlation to ground-truth quality. 

% It also comes close to or outperforms more costly approaches like supervised or ensemble-based QE in certain settings.

\end{abstract}

\section{Introduction}
Text generation models, such as transcription and translation systems like Whisper \cite{radford2023robust} or Large Language Models like Llama \cite{touvron2023llama}, have demonstrated remarkable effectiveness across various applications \cite{amorese2023automatic, xie2024me, masalkhi2024side}.
However, these models could still make mistakes in certain cases, such as when the input is noisy or when the context involves ambiguous phrasing or domain-specific jargon \cite{katkov2024benchmarking,huang2023survey}.
Consequently, it is crucial to inform users about the reliability of model outputs by offering a quality assessment. This task is formally recognized as Quality Estimation.

Particularly, Quality Estimation (QE) is the task of providing quality scores of model outputs during inference when the ground truth is not available. The most straightforward way is to utilize the model's output probability. 
While previous works have shown that model probability is prone to be overconfident \cite{nguyen2015deep,li2021confidence}, in this work, we point out another issue.
We show that the output probability on free-form text generation tasks, such as translation or summarization, can be \textbf{underconfident}. 
Specifically, lower probability does not necessarily indicate lower output quality, but could mean that the probability distribution is spread out over multiple correct options. 

We propose a simple QE approach, \textbf{\textsc{BoostedProb}}, which only utilizes the model output probability distribution. \textsc{BoostedProb} tackles the underconfidence phenomenon mentioned above by boosting the model's confidence scores when there are potentially multiple correct output options.

% in the cases where there are multiple tokens with dominating probability values in the model output distribution. 

Specifically, our contributions are as follows:

\begin{enumerate}
    \item We show that, for models performing free-form text generation tasks, at an output step,  there can be multiple valid outputs, leading to multiple tokens having dominant mass in the probability distribution. Probability mass spread over these valid tokens makes the model appear \textbf{underconfident}.
    
    \item We propose a QE approach, \textbf{\textsc{BoostedProb}},
    % \footnote{Code submitted as zip file.},
    that boosts the confidence of these dominant tokens. \textsc{BoostedProb} is easy to implement and does not add any complexity compared to raw model probabilities. It is substantially more efficient than ensemble-based QE, which requires generating multiple outputs, and supervised QE, which is data-dependent and not available for tasks other than translation.
    \item We show that \textsc{BoostedProb} is: (1) notably better as a quality estimator than the raw probabilities across different tasks and models; (2) coming close to or outperforming more expensive supervised and ensemble-based baselines in certain settings; (3) with \textsc{BoostedProb}, improving models' quality comes with improving their self-evaluation ability.

\end{enumerate}

% At each output step, our approach checks whether there are tokens with dominant probability mass in the probability distribution. If a dominant token is selected in the final output, the uncertainty score assigned to that output step would be the sum of the probability mass of all dominant tokens. If a non-dominant token is selected, we set the uncertainty score to zero. The uncertainty score of an output sequence would then be the average of the uncertainty score of each output step.

\section{Related Work}

\subsection{Quality Estimation} 
Model probability is the most trivial estimator of the output quality. However, previous works have shown that using the probability of the final output alone is not optimal, as neural models tend to be overconfident \cite{nguyen2015deep,li2021confidence}. 
Another approach is to use the entropy of the whole probability distribution \cite{fomicheva-etal-2020-unsupervised}. 
However, it does not consider which option is selected in the end. These methods are generally low-effort, with the only drawback that output probability might not be accessible for API-only models. Therefore, probability-based QE has been employed in many use cases, such as for deciding whether to ask users to repeat themselves in dialog systems \cite{jm3}, or determining the exit layer in early exiting models \cite{teerapittayanon2016branchynet, xin-etal-2020-deebert}.

Other types of QE are usually more costly. Some approaches require generating multiple outputs, such as ensemble-based approaches like Monte Carlo sequence entropy \cite{malinin2020uncertainty,kuhn2023semantic}, Perturbation-based QE \cite{dinh-niehues-2023-perturbation}, and self-validation approaches \cite{kadavath2022language}. 
Some approaches require access to the model training data to detect out-of-distribution instances during inference \cite{NEURIPS2018_abdeb6f5, ren2023outofdistribution}. 
Other approaches require an external model to measure the output quality. Prism \cite{thompson-post-2020-automatic} uses a multilingual Machine Translation model to score output from other models by forced decoding. \citet{cohen-etal-2023-lm} uses an examiner model to ask questions and discover inconsistencies of the evaluated model.

One outstanding case of external QE modules is supervised QE models for Machine Translation (MT), such as CometKiwi \cite{rei-etal-2022-cometkiwi}. For MT, there exists abundant data of \textit{(source, model translation, human-labeled scores)} tuples, which enables training supervised QE models. One can try to avoid the use of costly human-labeled scores by training QE models on synthetic data with synthetic errors \cite{tuan-etal-2021-quality}, or synthetic scores using reference-based metrics \cite{zouhar-etal-2023-poor} like BLEU \cite{papineni-etal-2002-bleu}, chrF \cite{popovic-2015-chrf} or BERTScore \cite{zhang2019bertscore}.
Supervised QE has been widely adopted in MT, and is getting close to the performance of reference-based metrics \cite{freitag-etal-2022-results}.

Nevertheless, supervised approaches are data-dependent, and mostly not available for tasks other than translation. Thus, we focus on using model probability as a quality estimator, given its simplicity and efficiency. Previous works mostly focus on the overconfidence problem of model probability, where one solution is to use larger models with more training data \cite{naganuma2023empirical,chhikara2025mind}. We identify another weakness of model probability - being \textbf{underconfident} for free-form text generation tasks, and propose a simple modification to the probability to tackle this.

\subsection{Dominant Tokens} \label{sec:sampling}
Previous works have considered that there can be multiple tokens with dominant probability mass in the output distribution. For example, \citet{ott2018analyzing} shows that, for MT, model distribution is highly spread in the hypothesis space. However, they focus on its effect on model fitting and inference search rather than on QE. Other works focus on sampling, where they try to find the set of dominant tokens to sample from during output generation to maintain high quality but also high diversity. Popular sampling strategies includes top-$k$ \cite{fan-etal-2018-hierarchical}, top-$p$ \cite{holtzmancurious}, $\epsilon$-cut \cite{hewitt-etal-2022-truncation}, $\eta$-cut \cite{hewitt-etal-2022-truncation} and min-$p$ \cite{nguyen2024turning}. For top-$k$, the assumption is that, the top $k$ tokens with the highest probability are the most important ones. For top-$p$, the most important tokens are those with top probabilities that sum up to $p$. 
For $\epsilon$-cut, the most important token probabilities are larger than $\epsilon$. 
For $\eta$-cut, the most important token probabilities are larger than either $\eta$ or $\sqrt{\eta} \times \exp(-entropy(\mathbb{P}))$, where $\mathbb{P}$ is the output probability distribution.
For min-$p$, the most important tokens have probabilities larger than the top-1 probability multiplied by $p$.

In our work, we focus on finding dominant tokens to boost their confidence for QE, rather than to support sampling or search strategies during inference.

\begin{figure*}[h]
    \centering
    \includegraphics[width=\linewidth]{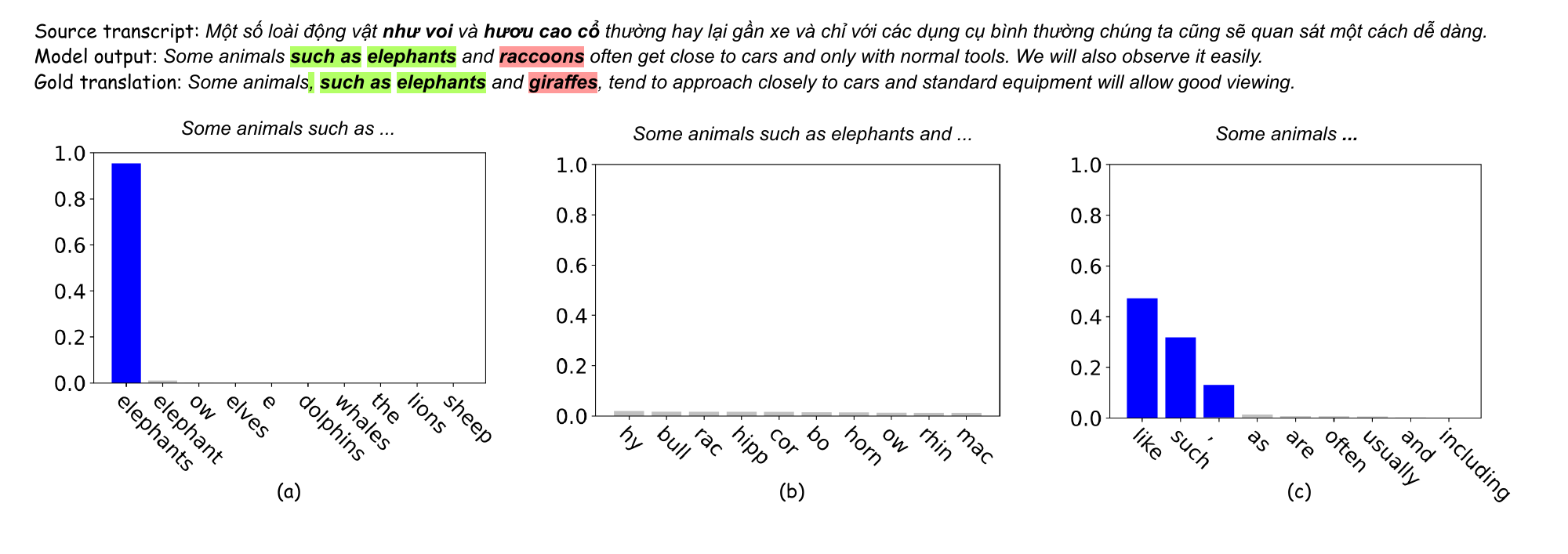}
    \caption{Whisper's output probability distributions. (a) The model gives high probability to the correct translation ("\textit{elephants}"). (b) The model gives low probability to all tokens, and outputs the wrong translation in the end ("\textit{raccoons}" instead of "\textit{giraffes}"). (c) The probabilities are lower due to probability mass being spread out between multiple correct options (",", "\textit{like}" and "\textit{such as}), and \textbf{\textit{do not indicate lower quality}}.}
    \label{fig:example}
\end{figure*}

\section{Model Probability -- An Underconfident Quality Estimator?} \label{sec:motivation}

\paragraph{Illustrative example} Our investigation begins with the example in Figure \ref{fig:example}, where Whisper Large V3 \cite{radford2023robust} translates a Vietnamese audio sentence into English. Figure \ref{fig:example}a (correct translation to "\textit{elephants}" and Figure \ref{fig:example}b (wrong translation to "\textit{raccoons}") are intuitive: higher probabilities indicate better output quality. However, in Figure \ref{fig:example}c, most probability mass is spread between three options:  the comma ",", "\textit{like}" and "\textit{such as}", all of which are reasonable outputs. The probabilities here are lower, but do not indicate low output quality. We suspect this happens due to the ambiguous nature of the Speech Translation task.

\paragraph{Ambiguous Tasks} By "ambiguous tasks", we refer to tasks where for an input, there can be multiple valid output options. We investigate the model behaviors when working on text-generation tasks with ambiguity like Speech Translation (ST), where for an input, multiple translations can be valid. We do so by comparing to the less ambiguous Automatic Speech Recognition (ASR) task, where for an input audio, there is only one correct transcription. The comparison is detailed below.

\begin{figure}[htbp]
    \centering
    \begin{subfigure}{0.15\textwidth}
        \centering
        \includegraphics[width=\linewidth]{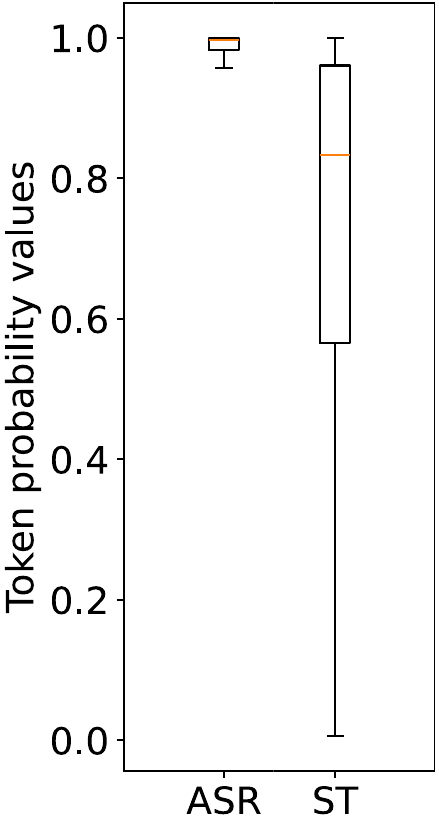}
        \caption{}
        \label{fig:asr_st_prob}
    \end{subfigure}
    \hfill
    \begin{subfigure}{0.147\textwidth}
        \centering
        \includegraphics[width=\linewidth]{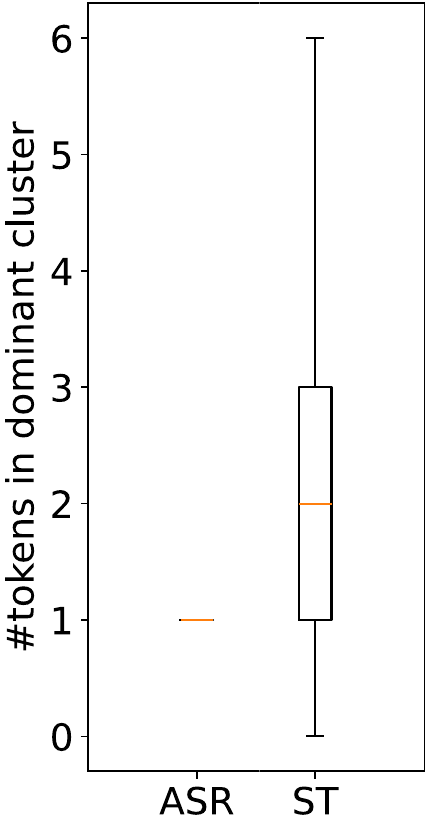}
        \caption{}
        \label{fig:asr_st_nr_dominant}
    \end{subfigure}
    \hfill
    \begin{subfigure}{0.143\textwidth}
        \centering
        \includegraphics[width=\linewidth]{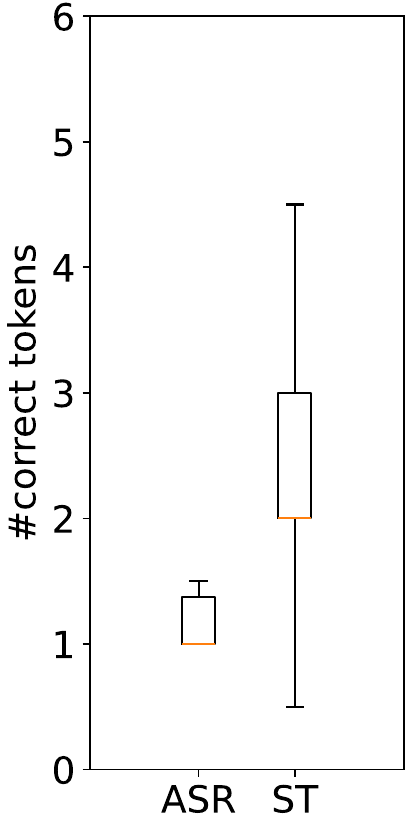}
        \caption{}
        \label{fig:asr_st_nr_correct}
    \end{subfigure}
    \caption{Model behaviours on ASR versus ST: (a) output token probability, (b) nr. dominant tokens, and (c) nr. correct tokens at every output step.
    % Outliers are removed from the plots for easier interpretation.
    }
    \label{fig:asr_vs_st}
\end{figure}

\paragraph{Output probability} We analyze the output probability distributions of Whisper on the ASR and ST tasks of the Fleurs data \cite{conneau2023fleurs} on 4 language pairs: Vietnamese-English, German-English, Spanish-English and Chinese-English. Looking at Figure \ref{fig:asr_st_prob}, for ASR, most finally chosen tokens have very high probability values that are close to 1. In contrast, for ST, the probability of the finally chosen tokens spreads out much more. 

\paragraph{Dominant Tokens} We take a closer look at the number of tokens that have notably higher probability mass in the distribution. We refer to the set of these tokens as \textit{dominant cluster}, and the tokens themselves as \textit{dominant tokens}. We identify them automatically using our heuristic later described in 
Section \ref{sec:method}, and report on the size of the clusters in Figure \ref{fig:asr_st_nr_dominant}. Observe that clusters with sizes larger than 1 only exist for the ambiguous ST task. 

\paragraph{Valid Output Tokens} We present these top tokens to human annotators, and ask them to annotate which tokens are valid output (details in Appendix \ref{sec:human}). Looking at Figure \ref{fig:asr_st_nr_correct}, at each output step, most of the time, there is only 1 correct output for ASR, but more than 1 for ST. This indicates that, the more spread-out probability distribution and the existence of dominant clusters with size larger than 1 are indeed due to the ambiguity of the ST task.

\paragraph{Underconfidence in Ambiguous Tasks} Our analysis shows that, text-generation tasks with ambiguity introduce aleatoric uncertainty, i.e., uncertainty coming from the data, which differs from epistemic uncertainty, i.e., uncertainty coming from the model's incompetence. Aleatoric uncertainty makes the model appear underconfident, as the probability mass is spread over multiple valid options. We discuss this \textit{underconfidence} phenomenon more formally with a theoretical analysis of the softmax function in Appendix \ref{app:theoretical}, where we show that there exists an upperbound of the probability scores assigned to every correct token at an output step, which is dependent on the number of correct tokens.
This observation brings us to a simple modification to the model probability to improve its effectiveness as a quality estimator, as detailed below.

\section{\textsc{BoostedProb}} \label{sec:method}
We propose \textbf{\textsc{BoostedProb}}, a Quality Estimation approach which boosts the confidence of the tokens in the dominant clusters. The overall idea is that, when the output token is dominant, instead of using its own probability as the quality score, we use the total probability mass of the dominant cluster. 

\paragraph{Finding Dominant Tokens} First, we identify which tokens are in the dominant cluster given the output distribution. Previous methods designed for sampling might mistakenly account for tokens with very low probability as dominant if they happen to, e.g., be in the top‑$k$ of the probability distribution, or fall within the top‑$p$ cumulative probability mass. For sampling, this might not be a big issue, since tokens with very low probability are unlikely to be selected anyway. However, for QE, it is problematic since we would mistakenly boost the confidence of low-quality output tokens. Therefore, we propose a heuristic that looks for the dominant tokens in a stricter manner. We look for a sudden drop in the sorted probability values in order to separate dominant from non-dominant tokens.  

\begin{figure}[h]
    \centering    
    \includegraphics[width=0.8\linewidth]{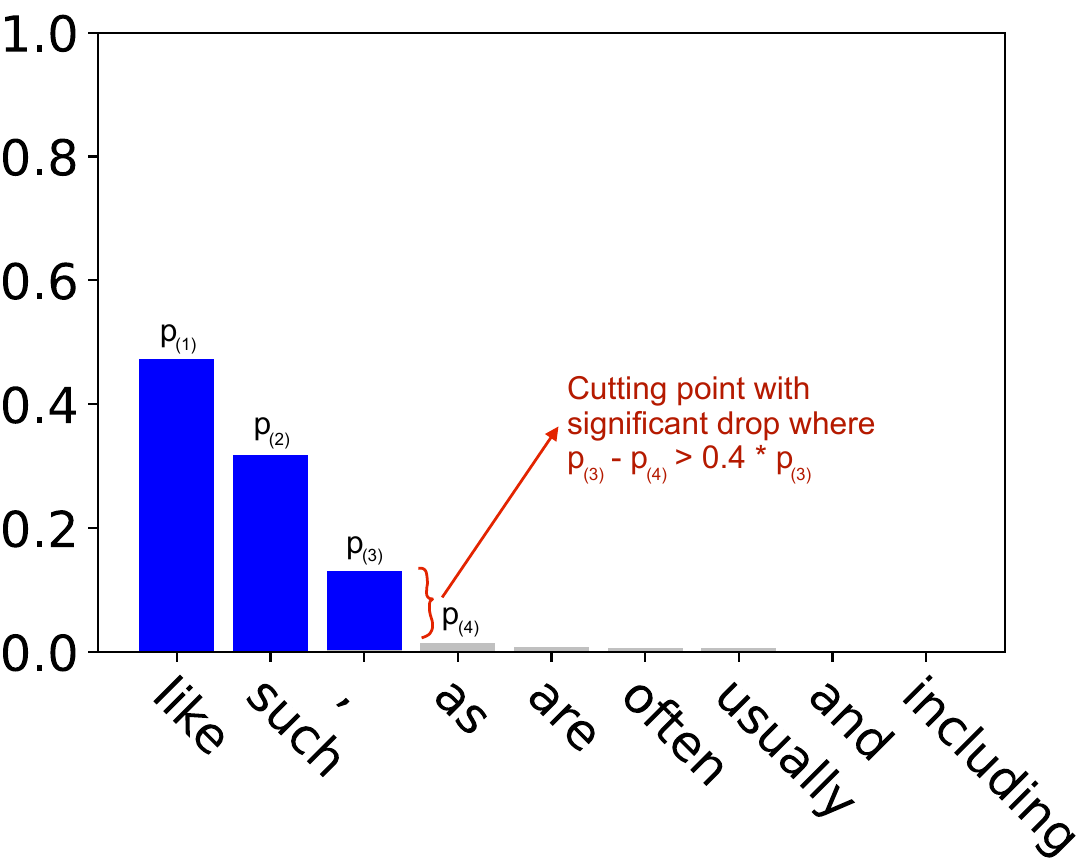}
    \caption{A dominant cluster found by our heuristic.}
    \label{fig:method}
\end{figure}

In particular: let $X=x_1,..,x_{|X|}$ be the input sequence, and $Y=y_1,..,y_{|Y|}$ be the model output. At an output step $t$, let the model probability distribution over the vocabulary $V$ be $\mathcal{P} = (p_1, p_2, \dots, p_{|V|})$, where $p_i = \mathbb{P}(y_t = w_i \mid y_{<t}, X)$ is the probability assigned to token $w_i$ at step $t$. First, we sort the probability distribution $\mathcal{P}$:
$$
\mathcal{P}_{\text{sorted}} = (p_{(1)}, p_{(2)}, \dots, p_{(|V|)}),
$$  
where \( p_{(1)} \geq p_{(2)} \geq \dots \geq p_{(|V|)} \). Then, we calculate the drops at each position, i.e., the differences between two consecutive probability values:

\begin{align*}
\mathcal{P}_{\text{diff}} & = \mathcal{P}_{\text{sorted}}  - \text{Shift}(\mathcal{P}_{\text{sorted}}) \\
 & = (p_{(1)}, p_{(2)}, \dots, p_{(|V|-1)}) \\
 & \hspace{2cm}- (p_{(2)}, p_{(3)}, \dots, p_{(|V|)})
\end{align*}

We then check at which positions the drops are significant. We propose a heuristic: if the drop is larger than $x\%$, then it is significant:
\begin{align*}
&\mathcal{P}_{\text{isSignificantDrop}} 
= \mathcal{P}_{\text{diff}} > \mathcal{P}_{\text{sorted}} \times x\% \\
&= (p_{(i)} - p_{(i+1)} > p_{(i)} \times x\% \text{\hspace{0.2cm}for } i = 1..|V|-1)
\end{align*}
Towards the distribution tail, the probabilities get close to zero, thus many drops satisfy the above condition although they are not significant drops that intuitively separate dominant from non-dominant tokens. Thus, we add another condition: the drop itself should be larger than a threshold $\epsilon$:
\begin{align*}
& \mathcal{P}_{\text{isSignificantDrop}} \\
&= (\mathcal{P}_{\text{diff}} > \mathcal{P}_{\text{sorted}} \times x\%) \textsc{ and } (\mathcal{P}_{\text{diff}} > \epsilon )  \\
&= (p_{(i)} - p_{(i+1)} > \max(p_{(i)} \times x\%, \epsilon)  \\
& \text{\hspace{0.6cm}for } i = 1..|V|-1)
\end{align*}

We arrived at a condition that considers both the relative value (drops larger than $x$\%) and absolute value (drops larger than $\epsilon$), making our approach more flexible in finding the dominant tokens in different probability distributions.

The last significant drop is then the cutting point:
$$
c = \max\{i \text{ | } \mathcal{P}_{\text{isSignificantDrop\_i}} = True\}
$$ 
where tokens with probabilities above the cutting point are dominant, and others are non-dominant. 
An illustration is shown in Figure \ref{fig:method}.

\paragraph{Extracting Token Quality Score} If the final output token is non-dominant, then we consider its own probability as the quality score.
If the finally selected token is dominant, we consider the total probability mass of the whole dominant cluster as the quality score. 
% That is, we take the sum of the probabilities of all dominant tokens. 
Particularly:
\begin{align*}
&QE(w_{(i)}) =
\begin{cases} 
p_{(i)}, & \text{if } i > c \\
\sum\limits_{j=1}^{c} p_{(j)}, & \text{otherwise } i \leq c \\
\end{cases}
\end{align*}
In this way, we favor the dominant tokens whose probability mass was spread amongst multiple sensible options, as described in Section \ref{sec:motivation}.

\paragraph{Extracting sequence quality estimation} The QE score for the output sequence $Y=y_1,..,y_{|Y|}$ is defined as the average of token-level QE scores:
$$
QE(Y) = \sum\limits_{t=1}^{|Y|} QE(y_t)
$$

We theoretically show that \textsc{BoostedProb} helps tackle the \textit{underconfidence} phenomenon discussed in Section \ref{sec:motivation} by allowing multiple correct output tokens to be assigned with a high, arbitrarily close to 1 score, which we detail in Appendix \ref{app:theoretical}.

\section{Experimental Setup} 
\hypertarget{task}{We test \textsc{BoostedProb} on different tasks: Speech Translation (ST), Machine/Text Translation (MT), Summarization (Sum.), Question Answering (QA).}

\subsection{Data} \label{sec:data}

The datasets are listed in Table \ref{tab:data}. All datasets contain the input and ground truth output. One exception is WMT22 General \cite{kocmi-etal-2022-findings}, which additionally contains candidate translations of participants in the WMT22 Shared Task, along with human-annotated quality scores (0 to 100) on the \textbf{segment level}. Another exception is HJQE \cite{yang-etal-2023-rethinking}, which additionally contains model translation output from the WMT20 QE Shared Task \cite{specia-etal-2020-findings-wmt} along with human-annotated quality labels (OK/BAD) on \textbf{token level}.

\begin{table}[htbp]
\small
\centering
\setlength{\tabcolsep}{4pt}
\begin{tabular}{llll}
Task \hyperlink{task}{*} & Dataset       & \#samples & Language      \\ \hline
ST   & Fleurs        & 350       & vi-en, de-en, \\
     & \cite{conneau2023fleurs}              &           & es-en, zh-en  \\
MT   & ParaCrawl     & 5000      & en-de, zh-en  \\
     & \cite{banon-etal-2020-paracrawl}              &           &               \\
     & WMT22 General & 2000      & en-de, zh-en  \\
     & \cite{kocmi-etal-2022-findings}              &           &               \\
     & HJQE          & 1000      & en-de, en-zh  \\
     & \cite{yang-etal-2023-rethinking}              &           &               \\
Sum. & XSum          & 3000      & en            \\
     & \cite{narayan-etal-2018-dont}              &           &               \\
QA   & GSM8k         & 3000      & en            \\
     & \cite{cobbe2021training}              &           &              \\
     & SciEx         & 1120      & en,de         \\
     & \cite{dinh-etal-2024-sciex}           &           &              
\end{tabular}
\caption{Data used in our experiments.}
\label{tab:data}
\end{table}

\subsection{Models}
The models used are listed in Table \ref{tab:model}. \textit{DeltaLM Large} is fine-tuned on 5M samples of ParaCrawl MT data, filtered by Bicleaner AI \cite{zaragoza-bernabeu-etal-2022-bicleaner,de-gibert-etal-2024-new}. Llama 3.3 70B is used with 4-bit quantization.

For smaller models, i.e., Whisper and DeltaLM, we generate output using beam search with beam size 4. For other models, we generate output with greedy search. 

\begin{table}[htbp]
\small
\centering
\setlength{\tabcolsep}{4pt}
\begin{tabular}{llr}
Task \hyperlink{task}{*} & Model              & Size \\ \hline
ST   & Whisper Large V3 \cite{radford2023robust}   & 1550M        \\
MT   & DeltaLM Large \cite{ma2021deltalm}   & 1374M        \\
     & NLLB \cite{costa2022no}              & 3.3B         \\
     & Tower \cite{tower_llm_2024}             & 7B           \\
Sum. & Bloomz \cite{muennighoff-etal-2023-crosslingual}             & 560M         \\
+ QA & Llama 3.2 \cite{touvron2023llama}         & 3B           \\
     & Llama 3.3 Instruct \cite{touvron2023llama} & 70B         
\end{tabular}
\caption{Models used in our experiments.}
\label{tab:model}
\end{table}

\subsection{Baselines}
\paragraph{Probability-based baselines} We consider \textit{raw model probability}, which uses the probability of the final output tokens, and \textit{probability entropy}, which uses the entropy of the whole probability distribution. These baselines are the most comparable to our approach, as they require only the probability distributions. We use them as the main baselines throughout our experiments. 

In some setups, we also consider more complex baselines, as detailed below.

\paragraph{Supervised QE Baseline} For some translation tasks, we use a supervised QE model, WMT22 CometKiwi DA \cite{rei-etal-2022-cometkiwi}. The model is trained on tuples of (\textsc{src}, \textsc{mt}, \textsc{da}), where \textsc{src} is the source sentence, \textsc{mt} is the MT output, and \textsc{da} is the Direct Assessment scores by humans. Note that this kind of supervised QE is mostly common for translation. For other tasks like summarization or question-answering, it is costly and not common to obtain such human-annotated quality. 
% We regard this as an upper baseline for our approach.

\begin{table*}[htbp]
\setlength{\tabcolsep}{4pt}
\small
\centering
\begin{tabular}{llllrrc}
                    & Model        & Test Set      & Language & \multicolumn{1}{l}{Probability} & \multicolumn{1}{l}{Entropy} & \multicolumn{1}{l}{\textsc{BoostedProb (Ours)}} \\ \hline
Speech Translation  & Whisper      & Fleurs        & vi-en    & 0.112                           & 0.379                       & \textbf{0.417}                         \\
                    &              & Fleurs        & de-en    & 0.213                           & \textbf{0.402}              & 0.385                                  \\
                    &              & Fleurs        & es-en    & 0.193                           & 0.295                       & \textbf{0.319}                         \\
                    &              & Fleurs        & cmn-en   & 0.053                           & 0.387                       & \textbf{0.424}                         \\ \hline
Machine Translation & DeltaLM      & WMT22 General & en-de    & 0.165                           & 0.169                       & \textbf{0.319}                         \\
                    &              & WMT22 General & zh-en    & 0.253                           & 0.082                       & \textbf{0.688}                         \\
                    & NLLB         & WMT22 General & en-de    & 0.141                           & 0.480                       & \textbf{0.525}                         \\
                    &              & WMT22 General & zh-en    & 0.182                           & 0.211                       & \textbf{0.289}                         \\
                    & Tower        & WMT22 General & en-de    & 0.158                           & 0.399                       & \textbf{0.414}                         \\
                    &              & WMT22 General & zh-en    & 0.005                           & 0.232                       & \textbf{0.240}                         \\ \hline
Summarization       & Bloomz 560M  & XSum          & en       & -0.003                          & 0.176                       & \textbf{0.210}                         \\
                    & Llama3.2 3B  & XSum          & en       & 0.002                           & 0.201                       & \textbf{0.209}                         \\
                    & Llama3.3 70B & XSum          & en       & 0.001                           & 0.000                       & \textbf{0.004}                         \\ \hline
Question Answering  & Bloomz 560M  & GSM8K         & en       & -0.002                          & \textbf{0.111}              & 0.009                                  \\
(Math)              & Llama3.2 3B  & GSM8K         & en       & -0.007                          & 0.006                       & \textbf{0.111}                         \\
                    & Llama3.3 70B & GSM8K         & en       & -0.001                          & 0.005                       & \textbf{0.006}                         \\ \hline
Question Answering  & Bloomz 560M  & SciEx         & en,de    & -0.002                          & 0.005                       & \textbf{0.006}                         \\
(University Exam)   & Llama3.2 3B  & SciEx         & en,de    & 0.002                           & 0.228                       & \textbf{0.310}                         \\
                    & Llama3.3 70B & SciEx         & en,de    & 0.103                           & \textbf{0.180}              & \textbf{0.180}                         \\ \hline
Average             &              &               &          & 0.083                           & 0.208                       & \textbf{0.268}                        
\end{tabular}
\caption{Performance of QE methods, in Pearson correlation to gold quality, across different tasks, models, test sets.}
\label{tab:overall}
\end{table*}

\paragraph{Unsupervised, Ensemble-based Baselines} For the word-level QE task on MT, we compare our approach with Perturbation-based QE \cite{dinh-niehues-2023-perturbation}, which makes minimal perturbations on the source input and measures the changes in the output as an indication of quality. For a subset of the experiments, we compare our approach with Monte Carlo sequence entropy \cite{malinin2020uncertainty,kuhn2023semantic}, which samples several output sequences and computes sequence-level entropy. These baselines are much more costly, as they require the generation of multiple outputs.

\paragraph{LLM self-judge} We also compare our approach against a recent baseline, LLM-as-a-Judge \cite{zheng2023judging}. To make this baseline more comparable to our reference-free QE, no-external-model setting, we adapt it to an LLM self-judge setup, where the model is asked to assign a quality score to its own output. This method requires an additional inference step and is not applicable to task-specific models such as Whisper and NLLB.

\subsection{Hyperparameters}
We loosely tune the hyperparameters, i.e., $x$ and $\epsilon$, on three models: Whisper, DeltaLM Large and Tower (see Appendix \ref{sec:hyperparamtune}). For these models, $x=30\%$ and $\epsilon=0.005$ are either the best or close to the best set of hyperparameters, showing that our approach is robust to hyperparameter setup. We then use these hyperparameters for all experiments.

\subsection{Evaluation}

On the segment level, we use Pearson correlation to measure how well QE methods correlate with the gold quality annotation. On the token level (HJQE dataset with OK/BAD labels), we use the Matthews correlation coefficient (MCC) scores \cite{matthews1975comparison}. The gold quality annotation is either automatically generated, or annotated by humans on pre-generated model output, as detailed below.

\subsubsection{Automatically Generated Gold Quality}
We create pseudo ground-truth quality scores to evaluate our reference-free QE methods using reference-based metrics. Reference-based metrics use human ground-truth answers in order to assign a quality score to a model output. We expect reference-based metrics to produce more reliable quality scores compared to reference-free QE methods, thus choosing them as pseudo ground-truth for evaluation.

\paragraph{Speech and Text Translation} We use XCOMET-XL \cite{guerreiro-etal-2024-xcomet} to generate pseudo ground-truth quality for translations. XCOMET-XL is a reference-based neural model. \citet{fixed_dinh-etal-2024-quality} showed that, for MT, such reference-based neural metrics are good enough to be used as the ground truth to rank reference-free QE metrics.

\paragraph{Summarization and Question Answering} We use BART Score \cite{NEURIPS2021_e4d2b6e6} as pseudo ground-truth output quality. The quality scores are calculated as the BART \cite{lewis-etal-2020-bart} model probability of the output given the input text. Unlike for MT, there has not been any study showing that reference-based metrics like Bart Score are sufficient as the ground truth for reference-free QE metrics. Therefore, we additionally report on other reference-based metrics in Appendix \ref{sec:more_ref}, including RougeL \cite{lin-2004-rouge}, BertScore \cite{zhang2019bertscore}, and LLM-as-a-Judge \cite{zheng2023judging} with Qwen2.5 72B Instruct \cite{qwen2.5}.

\subsubsection{Human-labeled Gold Quality} \label{sec:forced_decoding}
As described in Section \ref{sec:data}, the WMT22 General and the HJQE datasets contain human-annotated quality labels on pre-generated output. To utilize these labels, we use the translation models of consideration to re-generate the output presented in these datasets with forced decoding, also known as reference-free Prism \cite{thompson-post-2020-automatic}.

\section{Results and Discussion}
\subsection{Overall Performance} \label{sec:results_overall}

The overall performance of \textsc{BoostedProb}, in comparison with the \textit{raw probability} and \textit{probability entropy} baselines, is shown in Table \ref{tab:overall}. \textsc{BoostedProb} consistently outperforms \textit{raw probability} by a large margin (+0.194 Pearson correlation on average).
\textit{Probability entropy} appears to be a stronger baseline. This is expected since it takes into account the whole probability distribution at each output step. However, unlike \textsc{BoostedProb}, \textit{probability entropy} does not consider which token was finally selected. Therefore, \textsc{BoostedProb} on average still has better performance than \textit{probability entropy} (+0.065 in Pearson correlation).

The performance of \textsc{BoostedProb} is consistent for translation. It obtains more than 0.2 Pearson correlation across all settings. On the other hand, we observe cases where the two baselines fail. On Fleurs \textit{zh-en} with Whisper and WMT22 General \textit{zh-en} with Tower, \textit{raw probability} has very low performance, at 0.053 and 0.005 in Pearson correlation, respectively, while \textsc{BoostedProb} achieves 0.424 and 0.240. On WMT22 General \textit{zh-en} with DeltaLM, \textit{probability entropy} obtains 0.082 score, while \textsc{BoostedProb} achieves 0.688. This is possibly due to \textsc{BoostedProb} looking at both the whole probability distribution as well as which token is selected, differing from the two baselines.

The performance on Summarization and Question Answering is more inconsistent. In some settings, all methods have very low performance, under 0.1 in Pearson correlation. This could be due to the complexity of the task, or Bart Score gold quality labels are not sufficient to rank QE methods.

We also compare \textsc{BoostedProb} with more advanced methods, namely Monte Carlo sequence entropy \cite{malinin2020uncertainty,kuhn2023semantic} and LLM self-judge. As shown in Table~\ref{tab:monte}, the results are mixed: our approach outperforms these baselines in certain settings. It is important to note, however, that these baselines are more costly and less flexible. Monte Carlo sequence entropy requires generating multiple output samples, while LLM self-judge requires an additional inference pass and is not applicable to task-specific models.

\begin{table}[htbp]
\small
\setlength{\tabcolsep}{3.5pt}
\begin{tabular}{lllccc}
     & Model       & Lang. & Monte          & LLM Self           & Boosted        \\
     &             &       & Carlo          & judge          & Prob           \\ \hline
MT   & NLLB        & en-de & 0.303          & -              & \textbf{0.525} \\
     &             & zh-en & \textbf{0.337} & -              & 0.289          \\
     & Tower       & en-de & 0.302          & -              & \textbf{0.414} \\
     &             & zh-en & \textbf{0.240} & -              & \textbf{0.240} \\ \hline
Sum. & Bloomz 560M & en    & \textbf{0.236} & 0.149          & 0.210          \\
     & Llama3.2 3B & en    & 0.005          & 0.159          & \textbf{0.209} \\ \hline
QA   & Bloomz 560M & en    & 0.003          & \textbf{0.019} & 0.009          \\
     & Llama3.2 3B & en    & 0.175          & \textbf{0.232} & 0.111         
\end{tabular}
\caption{\textsc{BoostedProb} vs. Monte Carlo entropy and LLM self-judge. LLM self-judge is not applicable for task-specific models.$^1$}\label{tab:monte}
\end{table}
\footnotetext[1]{Tower is an LLM; however, we often observe that it fails in our specific self-judge setting, i.e., fails to produce a single quality score for a given translation. This is likely because the version of Tower we used was not trained for this task.}

One potential concern is whether the differences in the inference process affect the reported results. \textsc{BoostedProb} is designed to be applicable across inference methods (e.g., greedy decoding, beam search, top-p sampling), since it does not rely on the model always selecting the top-probability token. Furthermore, to address potential variability due to randomness, we provide example results across multiple runs with different random seeds in Appendix \ref{app:std}.

\subsection{Scoring Other Models' Output (Prism)} \label{sec:prism_results}
We evaluate \textsc{BoostedProb} on top of reference-free Prism: using MT models to score other translations with forced decoding. As scoring models, we use the model presented in the Prism paper, and NLLB, as it is a strong multilingual MT model, following the previous work of \citet{zouhar-etal-2024-fine}.

\begin{table}[htbp]
\small
\centering
\setlength{\tabcolsep}{4pt}
\begin{tabular}{lccccl}
Scoring Model               & \multicolumn{2}{c}{Prism Original}          & \multicolumn{2}{c}{NLLB}                    &                \\
Scored Model               & Best                 & Worst                & Best                 & Worst                &                \\
               & MT                   & MT                   & MT                   & MT                   & \textbf{Avg.}           \\ \hline
\textbf{en-de} & \multicolumn{1}{l}{} & \multicolumn{1}{l}{} & \multicolumn{1}{l}{} & \multicolumn{1}{l}{} &                \\
Probability    & 0.020                & 0.130                & 0.068                & 0.061                & 0.070          \\
Entropy        & 0.056                & 0.147                & 0.123                & 0.318                & 0.161          \\
\textsc{BoostedProb}    & 0.032                & 0.147                & 0.129                & 0.384                & {\ul 0.173}    \\
Supervised QE  & 0.202                & 0.453                & 0.202                & 0.453                & \textbf{0.328} \\ \hline
\textbf{zh-en} & \multicolumn{1}{l}{} & \multicolumn{1}{l}{} & \multicolumn{1}{l}{} & \multicolumn{1}{l}{} &                \\
Probability    & 0.205                & 0.285                & 0.153                & 0.085                & 0.182          \\
Entropy        & 0.246                & 0.299                & 0.095                & 0.194                & 0.209          \\
\textsc{BoostedProb}    & 0.251                & 0.317                & 0.153                & 0.231                & {\ul 0.238}    \\
Supervised QE  & 0.341                & 0.429                & 0.341                & 0.429                & \textbf{0.385}
\end{tabular}
\caption{Performance of QE methods with Prism, in Pearson correlation to human-labeled quality score.}
\label{tab:mt_fd}
\end{table}

% \begin{table}[h]
% \small
% \centering
% \begin{tabular}{lccc}
%                        & Best MT              & Worst MT             & \multicolumn{1}{l}{Average} \\ \hline
% {\ul \textbf{Scratch}} & \multicolumn{1}{l}{} & \multicolumn{1}{l}{} & \multicolumn{1}{l}{}        \\
% Probability            & 0.071                & 0.054                & 0.063                       \\
% Entropy                & 0.147                & 0.240                & 0.194                       \\
% Dominant               & 0.156                & 0.267                & 0.212                       \\ \hline
% {\ul \textbf{DeltaLM}} & \multicolumn{1}{l}{} & \multicolumn{1}{l}{} &                             \\
% Probability            & 0.070                & 0.064                & 0.067                       \\
% Entropy                & 0.161                & 0.308                & 0.235                       \\
% Dominant               & 0.178                & 0.338                & 0.258                       \\ \hline
% Supervised QE          & \textbf{0.202}                & \textbf{0.453}                & \textbf{0.328}                      
% \end{tabular}
% \caption{Performance of quality estimation methods, in Pearson correlation to human-labeled quality score}
% \label{tab:mt_fd}
% \end{table}

We make use of the WMT22 General Shared Task data. We select the best and the worst participation systems from the shared task, by taking the average of the human-labeled quality scores on all outputs of each system. We refer to them as \textit{Best MT} and \textit{Worst MT}. 
We calculate the correlation between the QE scores to the human-labeled score. 

From Table \ref{tab:mt_fd}, we can see that \textsc{BoostedProb} brings improvement on top of Prism, which originally made use of the raw model probability. As a result, it shrinks the gap between Prism and the more costly supervised QE baseline.
% without the need for human-labeled quality data.

\subsection{Word-level Quality Estimation} \label{sec:wl_qe}
We evaluate QE methods on annotating pre-generated translations with OK/BAD quality labels on HJQE. We again use Prism and NLLB as scoring models. We also use the original models that generated the translations in HJQE.
As the QE methods provide a continuous score, we use the development split of HJQE to find the best threshold to convert the scores to the OK/BAD labels.

\begin{table}[htbp]
\small
\centering
\begin{tabular}{lccc}
                              & en-de                & en-zh                & Avg.           \\ \hline
{\ul \textbf{NLLB}}           & \multicolumn{1}{l}{} &                      &                \\
Probability                   & 0.201                & 0.094                & 0.147          \\
Entropy                       & -0.042               & 0.007                & -0.017         \\
\textsc{BoostedProb}                   & {\ul 0.204}          & {\ul 0.123}          & {\ul 0.164}    \\ \hline
{\ul \textbf{Prism}}          & \multicolumn{1}{l}{} &                      &                \\
Probability                   & 0.157                & 0.084                & 0.120          \\
Entropy                       & -0.010               & 0.037                & 0.013          \\
\textsc{BoostedProb}                   & {\ul 0.193}          & {\ul 0.115}          & {\ul 0.154}    \\ \hline
{\ul \textbf{Original Model}} & \multicolumn{1}{l}{} & \multicolumn{1}{l}{} &                \\
Probability                   & 0.143                & 0.176                & 0.159          \\
Entropy                       & -0.074               & -0.142               & -0.108         \\
\textsc{BoostedProb}                   & {\ul 0.146}          & {\ul 0.232}          & {\ul 0.189}    \\ \hline
Perturbation-based QE         & 0.120                & 0.215                & 0.167          \\
Supervised QE                 & \textbf{0.220}       & \textbf{0.257}       & \textbf{0.239}
\end{tabular}
\caption{Performance of token-level QE in MCC scores.}
\label{tab:mcc}
\end{table}

The QE performance in MCC score is shown in Table \ref{tab:mcc}. We again observe that \textsc{BoostedProb} achieves the best performance among the probability-based QE methods, and comes closer to the performance of the supervised QE. In this experiment, we can see that the \textit{probability entropy} baseline fails. This is probably due to this baseline not considering the final output token. When evaluating on the sentence level, we hypothesize that the \textit{probability entropy} would at least indicate the quality of the model prefix during autoregressive generation, thus having reasonable performance, while failing completely in this case where each token is evaluated independently.

Using the original MT model, with \textsc{BoostedProb}, we outperform Perturbation-based QE on the \textit{en-zh} language pair. Note that with \textsc{BoostedProb}, we only need a single inference pass, unlike the Perturbation-based QE baseline.

\subsection{Effect of Generative Performance} \label{sec:fail}
\begin{figure}[htbp]
    \centering
    \includegraphics[width=0.9\linewidth]{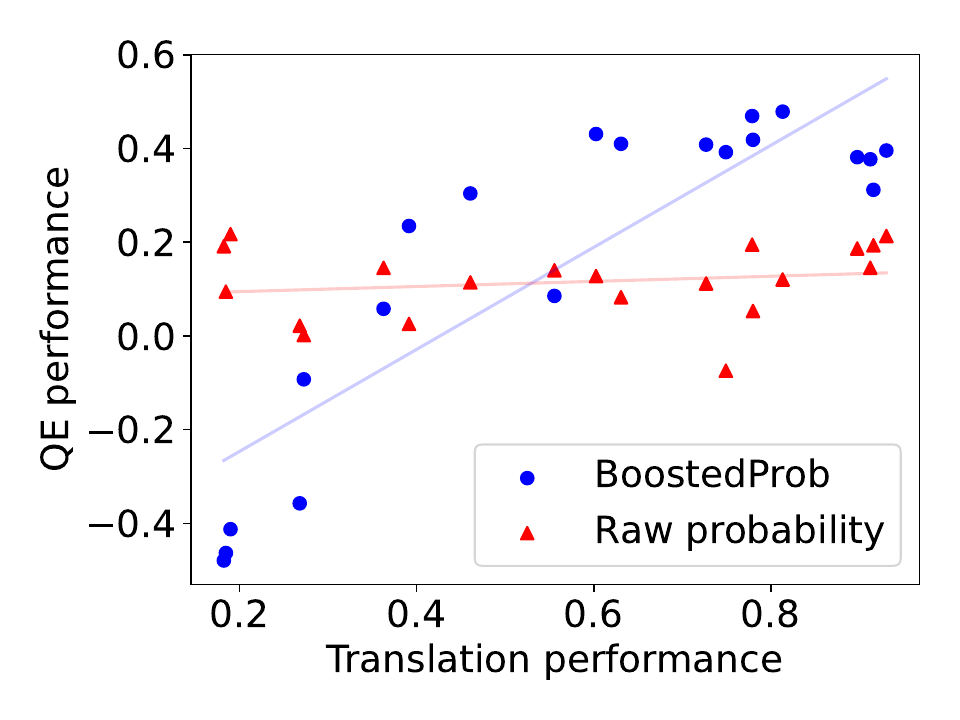}
    \caption{Model quality versus QE performance.}
    \label{fig:sys_performance_effect}
\end{figure}

We investigate how \textsc{BoostedProb} works for models of different quality. We investigate this on Speech Translation, with Whisper models of varying sizes for a more controlled experiment: Whisper Tiny, Whisper Base, Whisper Small, Whisper Medium, and Whisper Large V3. We run the models on the Fleurs test set on four different language pairs as before. We report the model translation performance (in XCOMET) and the QE performance (in Pearson) alongside in Figure \ref{fig:sys_performance_effect}. 

% The model performance is calculated as the average XCOMET score over all translation segments. The quality estimation performance is calculated as the Pearson correlation to the segment-level XCOMET scores, similar to before.

We see that the QE performance of the raw model probability stays consistently low. However, with \textsc{BoostedProb}, the QE performance improves as the quality of the model improves. This indicates that, with \textsc{BoostedProb}, improving a model's quality would come with improving its ability to do self-evaluation. Figure \ref{fig:sys_performance_effect} also exposes a limitation of our approach, since it worsens the QE performance compared to raw probability for very weak models. The reason might be due to our approach mistakenly emphasizing the weak models' overconfidence (Appendix \ref{sec:negative}). In these cases, we recommend using a stronger model to score the output of the weak model, rather than using the probability of the weak model to score itself, like in the Prism setting (Section \ref{sec:prism_results}).

\subsection{Finding Dominant Cluster}
\begin{table}[h]
\small
\centering 
\begin{tabular}{lllll}
                     & \multicolumn{2}{c}{MCC score}                                  & \multicolumn{2}{c}{Best hyperparams *}                \\
                     & \multicolumn{1}{c}{en-de}          & \multicolumn{1}{c}{en-zh} & \multicolumn{1}{c}{en-de} & \multicolumn{1}{c}{en-zh} \\ \hline
{\ul \textbf{Prism}} &                                    &                           &                           &                           \\
top-$k$                & \multicolumn{1}{c}{0.190}          & 0.107                     & $k$=7                       & $k$=7                       \\
$\epsilon$-cut          & \multicolumn{1}{c}{0.189}          & 0.112                     & $\epsilon$=0.01              & $\epsilon$=0.005             \\
$\eta$-cut              & \multicolumn{1}{c}{0.147}          & 0.100                     & $\epsilon$=0.005             & $\epsilon$=0.01              \\
top-$p$                & \multicolumn{1}{c}{0.119}          & 0.071                     & $p$=0.7                     & $p$=0.95                    \\
min-$p$                & \multicolumn{1}{c}{0.156}          & 0.085                     & $p$=0.95                    & $p$=0.90                    \\
jump-cut (ours)      & \multicolumn{1}{c}{\textbf{0.193}} & \textbf{0.115}            & $x$=0.3                     & $x$=0.3                     \\
                     &                                    &                           & $\epsilon$=0.005             & $\epsilon$=0.005             \\ \hline
{\ul \textbf{NLLB}}  &                                    &                           &                           &                           \\
top-$k$                & 0.203                              & \textbf{0.123}                     & $k$=5                       & $k$=5                       \\
$\epsilon$-cut          & 0.203                              & 0.121                     & $\epsilon$=0.05              & $\epsilon$=0.005             \\
$\eta$-cut              & 0.180                              & 0.105                     & $\epsilon$=0.2               & $\epsilon$=0.01              \\
top-$p$                & 0.171                              & 0.066                     & $p$=0.7                     & $p$=0.95                    \\
min-$p$                & 0.199                              & 0.095                     & $p$=0.7                     & $p$=0.80                    \\
jump-cut (ours)            & \textbf{0.204}                     & \textbf{0.123}                     & $x$=0.2                     & $x$=0.3                     \\
                     &                                    &                           & $\epsilon$=0.01              & $\epsilon$=0.005             \\ \hline
\multicolumn{5}{l}{* Best hyperparameters found on the dev split.}                                                                           
\end{tabular}
\caption{\textsc{BoostedProb}'s token-level QE performance when using different methods to find dominant clusters.}
\label{tab:diff_find}
\end{table}

We compare our method with other common methods, originally used for sampling (see Section \ref{sec:sampling} and \ref{sec:method}), to find the dominant tokens for \textsc{BoostedProb}. We again experiment on token-level QE on HJQE. We use the HJQE development split to find the best hyperparameter for each method.

The results are shown in Table \ref{tab:diff_find}. We denote our method as \textit{"jump-cut"}. Our method performs generally better than others, however, not by a large margin. Surprisingly, top-$k$ performs comparably to our approach despite naively assuming the dominant cluster size to be fixed. This might be due to two reasons. Firstly, the HJQE dev and test sets are potentially similar, thus tuning a good $k$ value is enough to achieve good performance. Secondly, as tokens with very low probability are unlikely to be chosen as the final output, it does not bring notable negative effect in terms of MCC score for top-$k$. 

However, as discussed in Section \ref{sec:method}, our method would still be a safer choice, as it would avoid mistakenly considering very low probability tokens as dominant, in the rare case that they are selected as the final output. Additionally, our method is more robust to hyperparameters: in three out of four settings in Table \ref{tab:diff_find}, we arrive at $x=30\%, \epsilon=0.005$, which are the same hyperparameter values we found before (Appendix \ref{sec:hyperparamtune}).

\section{Conclusion}
In this paper, we first perform automatic and manual analysis showing the existence of dominant clusters with sizes larger than 1 in model output probability distributions, which happens for ambiguous text-generation tasks. We show that the tokens in the dominant clusters are \textbf{underconfident}, as their probability is spread between multiple valid options. We proposed \textbf{\textsc{BoostedProb}} - a QE method that boosts the confidence of the dominant tokens. Since \textsc{BoostedProb} only utilizes the model probability distribution, it is low-cost, easy to implement, and can be applied to many model architectures. We show that \textsc{BoostedProb} performs notably better than model probability and probability entropy. It is also reaching close to or outperforming more costly approaches like supervised or ensemble-based QE in certain settings. With \textsc{BoostedProb}, improving models' quality comes with improving their self-evaluation ability.

\section{Limitations}
\textsc{BoostedProb} does not work well with very weak models, as it might emphasize weak models' overconfidence. This is mentioned in Section \ref{sec:fail}, and further discussed in Appendix \ref{sec:negative}. In these cases, we recommend using \textsc{BoostedProb} with a strong model to score the output of the weak model. This is the setting of Prism, which we have reported on in Section \ref{sec:prism_results}.
\textsc{BoostedProb} is also unlikely to have an effect for less ambiguous text generation tasks like Automatic Speech Recognition, or multiple-choice Question-Answering, since the dominant clusters with sizes larger than one are unlikely to appear (see Appendix \ref{sec:asr}). One can also argue that, unsupervised QE methods are in general not very useful whenever a supervised model exists (like in Speech Translation and Machine Translation). However, in these cases, our approach still brings benefits in terms of simplicity in implementation and inference time, compared to having an extra module for supervised QE in a translation pipeline.

\section*{Acknowledgements}
This work was supported by the Helmholtz Programme-oriented Funding, with project number 46.24.01, project name AI for Language Technologies. 
We acknowledge the HoreKa supercomputer funded by the Ministry of Science, Research and the Arts Baden-Wurttemberg and by the Federal Ministry of Education and Research. This work also received support from the European Union’s Horizon research and innovation programme under grant agreement No 101135798, project Meetween (My Personal AI Mediator for Virtual MEETtings BetWEEN People).

% Bibliography entries for the entire Anthology, followed by custom entries
\bibliography{anthology,custom}

\begin{thebibliography}{66}
\providecommand{\natexlab}[1]{#1}

\bibitem[{Alves et~al.(2024)Alves, Pombal, Guerreiro, Martins, Alves, Farajian, Peters, Rei, Fernandes, Agrawal, Colombo, de~Souza, and Martins}]{tower_llm_2024}
Duarte~M. Alves, José Pombal, Nuno~M. Guerreiro, Pedro~H. Martins, João Alves, Amin Farajian, Ben Peters, Ricardo Rei, Patrick Fernandes, Sweta Agrawal, Pierre Colombo, José G.~C. de~Souza, and André F.~T. Martins. 2024.
\newblock \href {https://arxiv.org/abs/2402.17733} {Tower: An open multilingual large language model for translation-related tasks}.
\newblock \emph{Preprint}, arXiv:2402.17733.

\bibitem[{Amorese et~al.(2023)Amorese, Greco, Cuciniello, Milo, Sheveleva, and Glackin}]{amorese2023automatic}
Terry Amorese, Claudia Greco, Marialucia Cuciniello, Rosa Milo, Olga Sheveleva, and Neil Glackin. 2023.
\newblock Automatic speech recognition (asr) with whisper: Testing performances in different languages.
\newblock In \emph{S3C@ CHItaly}, pages 1--8.

\bibitem[{Ba{\~n}{\'o}n et~al.(2020)Ba{\~n}{\'o}n, Chen, Haddow, Heafield, Hoang, Espl{\`a}-Gomis, Forcada, Kamran, Kirefu, Koehn, Ortiz~Rojas, Pla~Sempere, Ram{\'i}rez-S{\'a}nchez, Sarr{\'i}as, Strelec, Thompson, Waites, Wiggins, and Zaragoza}]{banon-etal-2020-paracrawl}
Marta Ba{\~n}{\'o}n, Pinzhen Chen, Barry Haddow, Kenneth Heafield, Hieu Hoang, Miquel Espl{\`a}-Gomis, Mikel~L. Forcada, Amir Kamran, Faheem Kirefu, Philipp Koehn, Sergio Ortiz~Rojas, Leopoldo Pla~Sempere, Gema Ram{\'i}rez-S{\'a}nchez, Elsa Sarr{\'i}as, Marek Strelec, Brian Thompson, William Waites, Dion Wiggins, and Jaume Zaragoza. 2020.
\newblock \href {https://doi.org/10.18653/v1/2020.acl-main.417} {{P}ara{C}rawl: Web-scale acquisition of parallel corpora}.
\newblock In \emph{Proceedings of the 58th Annual Meeting of the Association for Computational Linguistics}, pages 4555--4567, Online. Association for Computational Linguistics.

\bibitem[{Chhikara(2025)}]{chhikara2025mind}
Prateek Chhikara. 2025.
\newblock Mind the confidence gap: Overconfidence, calibration, and distractor effects in large language models.
\newblock \emph{arXiv preprint arXiv:2502.11028}.

\bibitem[{Cobbe et~al.(2021)Cobbe, Kosaraju, Bavarian, Chen, Jun, Kaiser, Plappert, Tworek, Hilton, Nakano et~al.}]{cobbe2021training}
Karl Cobbe, Vineet Kosaraju, Mohammad Bavarian, Mark Chen, Heewoo Jun, Lukasz Kaiser, Matthias Plappert, Jerry Tworek, Jacob Hilton, Reiichiro Nakano, et~al. 2021.
\newblock Training verifiers to solve math word problems.
\newblock \emph{arXiv preprint arXiv:2110.14168}.

\bibitem[{Cohen et~al.(2023)Cohen, Hamri, Geva, and Globerson}]{cohen-etal-2023-lm}
Roi Cohen, May Hamri, Mor Geva, and Amir Globerson. 2023.
\newblock \href {https://doi.org/10.18653/v1/2023.emnlp-main.778} {{LM} vs {LM}: Detecting factual errors via cross examination}.
\newblock In \emph{Proceedings of the 2023 Conference on Empirical Methods in Natural Language Processing}, pages 12621--12640, Singapore. Association for Computational Linguistics.

\bibitem[{Conneau et~al.(2023)Conneau, Ma, Khanuja, Zhang, Axelrod, Dalmia, Riesa, Rivera, and Bapna}]{conneau2023fleurs}
Alexis Conneau, Min Ma, Simran Khanuja, Yu~Zhang, Vera Axelrod, Siddharth Dalmia, Jason Riesa, Clara Rivera, and Ankur Bapna. 2023.
\newblock Fleurs: Few-shot learning evaluation of universal representations of speech.
\newblock In \emph{2022 IEEE Spoken Language Technology Workshop (SLT)}, pages 798--805. IEEE.

\bibitem[{Costa-Juss{\`a} et~al.(2022)Costa-Juss{\`a}, Cross, {\c{C}}elebi, Elbayad, Heafield, Heffernan, Kalbassi, Lam, Licht, Maillard et~al.}]{costa2022no}
Marta~R Costa-Juss{\`a}, James Cross, Onur {\c{C}}elebi, Maha Elbayad, Kenneth Heafield, Kevin Heffernan, Elahe Kalbassi, Janice Lam, Daniel Licht, Jean Maillard, et~al. 2022.
\newblock No language left behind: Scaling human-centered machine translation.
\newblock \emph{arXiv preprint arXiv:2207.04672}.

\bibitem[{de~Gibert et~al.(2024)de~Gibert, Nail, Arefyev, Ba{\~n}{\'o}n, van~der Linde, Ji, Zaragoza-Bernabeu, Aulamo, Ram{\'i}rez-S{\'a}nchez, Kutuzov, Pyysalo, Oepen, and Tiedemann}]{de-gibert-etal-2024-new}
Ona de~Gibert, Graeme Nail, Nikolay Arefyev, Marta Ba{\~n}{\'o}n, Jelmer van~der Linde, Shaoxiong Ji, Jaume Zaragoza-Bernabeu, Mikko Aulamo, Gema Ram{\'i}rez-S{\'a}nchez, Andrey Kutuzov, Sampo Pyysalo, Stephan Oepen, and J{\"o}rg Tiedemann. 2024.
\newblock \href {https://aclanthology.org/2024.lrec-main.100/} {A new massive multilingual dataset for high-performance language technologies}.
\newblock In \emph{Proceedings of the 2024 Joint International Conference on Computational Linguistics, Language Resources and Evaluation (LREC-COLING 2024)}, pages 1116--1128, Torino, Italia. ELRA and ICCL.

\bibitem[{Deoghare et~al.(2023)Deoghare, Kanojia, Blain, Ranasinghe, and Bhattacharyya}]{deoghare-etal-2023-quality}
Sourabh Deoghare, Diptesh Kanojia, Fred Blain, Tharindu Ranasinghe, and Pushpak Bhattacharyya. 2023.
\newblock \href {https://doi.org/10.18653/v1/2023.findings-emnlp.115} {Quality estimation-assisted automatic post-editing}.
\newblock In \emph{Findings of the Association for Computational Linguistics: EMNLP 2023}, pages 1686--1698, Singapore. Association for Computational Linguistics.

\bibitem[{Dinh et~al.(2024{\natexlab{a}})Dinh, Mullov, B{\"a}rmann, Li, Liu, Rei{\ss}, Lee, Lerzer, Gao, Peller-Konrad, R{\"o}ddiger, Waibel, Asfour, Beigl, Stiefelhagen, Dachsbacher, B{\"o}hm, and Niehues}]{dinh-etal-2024-sciex}
Tu~Anh Dinh, Carlos Mullov, Leonard B{\"a}rmann, Zhaolin Li, Danni Liu, Simon Rei{\ss}, Jueun Lee, Nathan Lerzer, Jianfeng Gao, Fabian Peller-Konrad, Tobias R{\"o}ddiger, Alexander Waibel, Tamim Asfour, Michael Beigl, Rainer Stiefelhagen, Carsten Dachsbacher, Klemens B{\"o}hm, and Jan Niehues. 2024{\natexlab{a}}.
\newblock \href {https://doi.org/10.18653/v1/2024.emnlp-main.647} {{S}ci{E}x: Benchmarking large language models on scientific exams with human expert grading and automatic grading}.
\newblock In \emph{Proceedings of the 2024 Conference on Empirical Methods in Natural Language Processing}, pages 11592--11610, Miami, Florida, USA. Association for Computational Linguistics.

\bibitem[{Dinh and Niehues(2023)}]{dinh-niehues-2023-perturbation}
Tu~Anh Dinh and Jan Niehues. 2023.
\newblock \href {https://aclanthology.org/2023.mtsummit-research.6/} {Perturbation-based {QE}: An explainable, unsupervised word-level quality estimation method for blackbox machine translation}.
\newblock In \emph{Proceedings of Machine Translation Summit XIX, Vol. 1: Research Track}, pages 59--71, Macau SAR, China. Asia-Pacific Association for Machine Translation.

\bibitem[{Dinh et~al.(2024{\natexlab{b}})Dinh, Palzer, and Niehues}]{fixed_dinh-etal-2024-quality}
Tu~Anh Dinh, Tobias Palzer, and Jan Niehues. 2024{\natexlab{b}}.
\newblock \href {https://aclanthology.org/2024.eamt-1.14/} {Quality estimation with $k$-nearest neighbors and automatic evaluation for model-specific quality estimation}.
\newblock In \emph{Proceedings of the 25th Annual Conference of the European Association for Machine Translation (Volume 1)}, pages 133--146, Sheffield, UK. European Association for Machine Translation (EAMT).

\bibitem[{Fadeeva et~al.(2023)Fadeeva, Vashurin, Tsvigun, Vazhentsev, Petrakov, Fedyanin, Vasilev, Goncharova, Panchenko, Panov, Baldwin, and Shelmanov}]{fadeeva-etal-2023-lm}
Ekaterina Fadeeva, Roman Vashurin, Akim Tsvigun, Artem Vazhentsev, Sergey Petrakov, Kirill Fedyanin, Daniil Vasilev, Elizaveta Goncharova, Alexander Panchenko, Maxim Panov, Timothy Baldwin, and Artem Shelmanov. 2023.
\newblock \href {https://doi.org/10.18653/v1/2023.emnlp-demo.41} {{LM}-polygraph: Uncertainty estimation for language models}.
\newblock In \emph{Proceedings of the 2023 Conference on Empirical Methods in Natural Language Processing: System Demonstrations}, pages 446--461, Singapore. Association for Computational Linguistics.

\bibitem[{Fan et~al.(2018)Fan, Lewis, and Dauphin}]{fan-etal-2018-hierarchical}
Angela Fan, Mike Lewis, and Yann Dauphin. 2018.
\newblock \href {https://doi.org/10.18653/v1/P18-1082} {Hierarchical neural story generation}.
\newblock In \emph{Proceedings of the 56th Annual Meeting of the Association for Computational Linguistics (Volume 1: Long Papers)}, pages 889--898, Melbourne, Australia. Association for Computational Linguistics.

\bibitem[{Fernandes et~al.(2022)Fernandes, Farinhas, Rei, C.~de Souza, Ogayo, Neubig, and Martins}]{fernandes-etal-2022-quality}
Patrick Fernandes, Ant{\'o}nio Farinhas, Ricardo Rei, Jos{\'e}~G. C.~de Souza, Perez Ogayo, Graham Neubig, and Andre Martins. 2022.
\newblock \href {https://doi.org/10.18653/v1/2022.naacl-main.100} {Quality-aware decoding for neural machine translation}.
\newblock In \emph{Proceedings of the 2022 Conference of the North American Chapter of the Association for Computational Linguistics: Human Language Technologies}, pages 1396--1412, Seattle, United States. Association for Computational Linguistics.

\bibitem[{Fomicheva et~al.(2020)Fomicheva, Sun, Yankovskaya, Blain, Guzm{\'a}n, Fishel, Aletras, Chaudhary, and Specia}]{fomicheva-etal-2020-unsupervised}
Marina Fomicheva, Shuo Sun, Lisa Yankovskaya, Fr{\'e}d{\'e}ric Blain, Francisco Guzm{\'a}n, Mark Fishel, Nikolaos Aletras, Vishrav Chaudhary, and Lucia Specia. 2020.
\newblock \href {https://doi.org/10.1162/tacl_a_00330} {Unsupervised quality estimation for neural machine translation}.
\newblock \emph{Transactions of the Association for Computational Linguistics}, 8:539--555.

\bibitem[{Freitag et~al.(2022)Freitag, Rei, Mathur, Lo, Stewart, Avramidis, Kocmi, Foster, Lavie, and Martins}]{freitag-etal-2022-results}
Markus Freitag, Ricardo Rei, Nitika Mathur, Chi-kiu Lo, Craig Stewart, Eleftherios Avramidis, Tom Kocmi, George Foster, Alon Lavie, and Andr{\'e} F.~T. Martins. 2022.
\newblock \href {https://aclanthology.org/2022.wmt-1.2/} {Results of {WMT}22 metrics shared task: Stop using {BLEU} {--} neural metrics are better and more robust}.
\newblock In \emph{Proceedings of the Seventh Conference on Machine Translation (WMT)}, pages 46--68, Abu Dhabi, United Arab Emirates (Hybrid). Association for Computational Linguistics.

\bibitem[{Guerreiro et~al.(2024)Guerreiro, Rei, Stigt, Coheur, Colombo, and Martins}]{guerreiro-etal-2024-xcomet}
Nuno~M. Guerreiro, Ricardo Rei, Daan~van Stigt, Luisa Coheur, Pierre Colombo, and Andr{\'e} F.~T. Martins. 2024.
\newblock \href {https://doi.org/10.1162/tacl_a_00683} {xcomet: Transparent machine translation evaluation through fine-grained error detection}.
\newblock \emph{Transactions of the Association for Computational Linguistics}, 12:979--995.

\bibitem[{Hewitt et~al.(2022)Hewitt, Manning, and Liang}]{hewitt-etal-2022-truncation}
John Hewitt, Christopher Manning, and Percy Liang. 2022.
\newblock \href {https://doi.org/10.18653/v1/2022.findings-emnlp.249} {Truncation sampling as language model desmoothing}.
\newblock In \emph{Findings of the Association for Computational Linguistics: EMNLP 2022}, pages 3414--3427, Abu Dhabi, United Arab Emirates. Association for Computational Linguistics.

\bibitem[{Holtzman et~al.()Holtzman, Buys, Du, Forbes, and Choi}]{holtzmancurious}
Ari Holtzman, Jan Buys, Li~Du, Maxwell Forbes, and Yejin Choi.
\newblock The curious case of neural text degeneration.
\newblock In \emph{International Conference on Learning Representations}.

\bibitem[{Huang et~al.(2023)Huang, Yu, Ma, Zhong, Feng, Wang, Chen, Peng, Feng, Qin et~al.}]{huang2023survey}
Lei Huang, Weijiang Yu, Weitao Ma, Weihong Zhong, Zhangyin Feng, Haotian Wang, Qianglong Chen, Weihua Peng, Xiaocheng Feng, Bing Qin, et~al. 2023.
\newblock A survey on hallucination in large language models: Principles, taxonomy, challenges, and open questions.
\newblock \emph{arXiv preprint arXiv:2311.05232}.

\bibitem[{Jurafsky and Martin(2025)}]{jm3}
Daniel Jurafsky and James~H. Martin. 2025.
\newblock \href {https://web.stanford.edu/~jurafsky/slp3/} {\emph{Speech and Language Processing: An Introduction to Natural Language Processing, Computational Linguistics, and Speech Recognition with Language Models}}, 3rd edition.
\newblock Online manuscript released January 12, 2025.

\bibitem[{Kadavath et~al.(2022)Kadavath, Conerly, Askell, Henighan, Drain, Perez, Schiefer, Hatfield-Dodds, DasSarma, Tran-Johnson et~al.}]{kadavath2022language}
Saurav Kadavath, Tom Conerly, Amanda Askell, Tom Henighan, Dawn Drain, Ethan Perez, Nicholas Schiefer, Zac Hatfield-Dodds, Nova DasSarma, Eli Tran-Johnson, et~al. 2022.
\newblock Language models (mostly) know what they know.
\newblock \emph{arXiv preprint arXiv:2207.05221}.

\bibitem[{Katkov et~al.(2024)Katkov, Liotta, and Vietti}]{katkov2024benchmarking}
Sergei Katkov, Antonio Liotta, and Alessandro Vietti. 2024.
\newblock Benchmarking whisper under diverse audio transformations and real-time constraints.
\newblock In \emph{International Conference on Speech and Computer}, pages 82--91. Springer.

\bibitem[{Kocmi et~al.(2022)Kocmi, Bawden, Bojar, Dvorkovich, Federmann, Fishel, Gowda, Graham, Grundkiewicz, Haddow, Knowles, Koehn, Monz, Morishita, Nagata, Nakazawa, Nov{\'a}k, Popel, and Popovi{\'c}}]{kocmi-etal-2022-findings}
Tom Kocmi, Rachel Bawden, Ond{\v{r}}ej Bojar, Anton Dvorkovich, Christian Federmann, Mark Fishel, Thamme Gowda, Yvette Graham, Roman Grundkiewicz, Barry Haddow, Rebecca Knowles, Philipp Koehn, Christof Monz, Makoto Morishita, Masaaki Nagata, Toshiaki Nakazawa, Michal Nov{\'a}k, Martin Popel, and Maja Popovi{\'c}. 2022.
\newblock \href {https://aclanthology.org/2022.wmt-1.1/} {Findings of the 2022 conference on machine translation ({WMT}22)}.
\newblock In \emph{Proceedings of the Seventh Conference on Machine Translation (WMT)}, pages 1--45, Abu Dhabi, United Arab Emirates (Hybrid). Association for Computational Linguistics.

\bibitem[{Kuhn et~al.(2023)Kuhn, Gal, and Farquhar}]{kuhn2023semantic}
Lorenz Kuhn, Yarin Gal, and Sebastian Farquhar. 2023.
\newblock Semantic uncertainty: Linguistic invariances for uncertainty estimation in natural language generation.
\newblock \emph{International Conference on Learning Representations, ICLR}.

\bibitem[{Lee et~al.(2018)Lee, Lee, Lee, and Shin}]{NEURIPS2018_abdeb6f5}
Kimin Lee, Kibok Lee, Honglak Lee, and Jinwoo Shin. 2018.
\newblock \href {https://proceedings.neurips.cc/paper_files/paper/2018/file/abdeb6f575ac5c6676b747bca8d09cc2-Paper.pdf} {A simple unified framework for detecting out-of-distribution samples and adversarial attacks}.
\newblock In \emph{Advances in Neural Information Processing Systems}, volume~31. Curran Associates, Inc.

\bibitem[{Lewis et~al.(2020)Lewis, Liu, Goyal, Ghazvininejad, Mohamed, Levy, Stoyanov, and Zettlemoyer}]{lewis-etal-2020-bart}
Mike Lewis, Yinhan Liu, Naman Goyal, Marjan Ghazvininejad, Abdelrahman Mohamed, Omer Levy, Veselin Stoyanov, and Luke Zettlemoyer. 2020.
\newblock \href {https://doi.org/10.18653/v1/2020.acl-main.703} {{BART}: Denoising sequence-to-sequence pre-training for natural language generation, translation, and comprehension}.
\newblock In \emph{Proceedings of the 58th Annual Meeting of the Association for Computational Linguistics}, pages 7871--7880, Online. Association for Computational Linguistics.

\bibitem[{Li et~al.(2021)Li, Qiu, Zhang, Li, He, Woodland, Cao, and Strohman}]{li2021confidence}
Qiujia Li, David Qiu, Yu~Zhang, Bo~Li, Yanzhang He, Philip~C Woodland, Liangliang Cao, and Trevor Strohman. 2021.
\newblock Confidence estimation for attention-based sequence-to-sequence models for speech recognition.
\newblock In \emph{ICASSP 2021-2021 IEEE International Conference on Acoustics, Speech and Signal Processing (ICASSP)}, pages 6388--6392. IEEE.

\bibitem[{Lin(2004)}]{lin-2004-rouge}
Chin-Yew Lin. 2004.
\newblock \href {https://aclanthology.org/W04-1013/} {{ROUGE}: A package for automatic evaluation of summaries}.
\newblock In \emph{Text Summarization Branches Out}, pages 74--81, Barcelona, Spain. Association for Computational Linguistics.

\bibitem[{Lin and Och(2004)}]{lin-och-2004-automatic}
Chin-Yew Lin and Franz~Josef Och. 2004.
\newblock \href {https://doi.org/10.3115/1218955.1219032} {Automatic evaluation of machine translation quality using longest common subsequence and skip-bigram statistics}.
\newblock In \emph{Proceedings of the 42nd Annual Meeting of the Association for Computational Linguistics ({ACL}-04)}, pages 605--612, Barcelona, Spain.

\bibitem[{Ma et~al.(2021)Ma, Dong, Huang, Zhang, Muzio, Singhal, Awadalla, Song, and Wei}]{ma2021deltalm}
Shuming Ma, Li~Dong, Shaohan Huang, Dongdong Zhang, Alexandre Muzio, Saksham Singhal, Hany~Hassan Awadalla, Xia Song, and Furu Wei. 2021.
\newblock Deltalm: Encoder-decoder pre-training for language generation and translation by augmenting pretrained multilingual encoders.
\newblock \emph{arXiv preprint arXiv:2106.13736}.

\bibitem[{Malinin and Gales(2021)}]{malinin2020uncertainty}
Andrey Malinin and Mark Gales. 2021.
\newblock Uncertainty estimation in autoregressive structured prediction.
\newblock \emph{International Conference on Learning Representations, ICLR}.

\bibitem[{Masalkhi et~al.(2024)Masalkhi, Ong, Waisberg, Zaman, Sarker, Lee, and Tavakkoli}]{masalkhi2024side}
Mouayad Masalkhi, Joshua Ong, Ethan Waisberg, Nasif Zaman, Prithul Sarker, Andrew~G Lee, and Alireza Tavakkoli. 2024.
\newblock A side-by-side evaluation of llama 2 by meta with chatgpt and its application in ophthalmology.
\newblock \emph{Eye}, pages 1--4.

\bibitem[{Matthews(1975)}]{matthews1975comparison}
Brian~W Matthews. 1975.
\newblock Comparison of the predicted and observed secondary structure of t4 phage lysozyme.
\newblock \emph{Biochimica et Biophysica Acta (BBA)-Protein Structure}, 405(2):442--451.

\bibitem[{Muennighoff et~al.(2023)Muennighoff, Wang, Sutawika, Roberts, Biderman, Le~Scao, Bari, Shen, Yong, Schoelkopf, Tang, Radev, Aji, Almubarak, Albanie, Alyafeai, Webson, Raff, and Raffel}]{muennighoff-etal-2023-crosslingual}
Niklas Muennighoff, Thomas Wang, Lintang Sutawika, Adam Roberts, Stella Biderman, Teven Le~Scao, M~Saiful Bari, Sheng Shen, Zheng~Xin Yong, Hailey Schoelkopf, Xiangru Tang, Dragomir Radev, Alham~Fikri Aji, Khalid Almubarak, Samuel Albanie, Zaid Alyafeai, Albert Webson, Edward Raff, and Colin Raffel. 2023.
\newblock \href {https://doi.org/10.18653/v1/2023.acl-long.891} {Crosslingual generalization through multitask finetuning}.
\newblock In \emph{Proceedings of the 61st Annual Meeting of the Association for Computational Linguistics (Volume 1: Long Papers)}, pages 15991--16111, Toronto, Canada. Association for Computational Linguistics.

\bibitem[{Naganuma et~al.(2025)Naganuma, Hataya, and Mitliagkas}]{naganuma2023empirical}
Hiroki Naganuma, Ryuichiro Hataya, and Ioannis Mitliagkas. 2025.
\newblock An empirical study of pre-trained model selection for out-of-distribution generalization and calibration.
\newblock \emph{Transactions on Machine Learning Research}.

\bibitem[{Narayan et~al.(2018)Narayan, Cohen, and Lapata}]{narayan-etal-2018-dont}
Shashi Narayan, Shay~B. Cohen, and Mirella Lapata. 2018.
\newblock \href {https://doi.org/10.18653/v1/D18-1206} {Don`t give me the details, just the summary! topic-aware convolutional neural networks for extreme summarization}.
\newblock In \emph{Proceedings of the 2018 Conference on Empirical Methods in Natural Language Processing}, pages 1797--1807, Brussels, Belgium. Association for Computational Linguistics.

\bibitem[{Nguyen et~al.(2015)Nguyen, Yosinski, and Clune}]{nguyen2015deep}
Anh Nguyen, Jason Yosinski, and Jeff Clune. 2015.
\newblock Deep neural networks are easily fooled: High confidence predictions for unrecognizable images.
\newblock In \emph{Proceedings of the IEEE conference on computer vision and pattern recognition}, pages 427--436.

\bibitem[{Nguyen et~al.(2024)Nguyen, Baker, Neo, Roush, Kirsch, and Shwartz-Ziv}]{nguyen2024turning}
Minh Nguyen, Andrew Baker, Clement Neo, Allen Roush, Andreas Kirsch, and Ravid Shwartz-Ziv. 2024.
\newblock Turning up the heat: Min-p sampling for creative and coherent llm outputs.
\newblock \emph{arXiv preprint arXiv:2407.01082}.

\bibitem[{Ott et~al.(2018)Ott, Auli, Grangier, and Ranzato}]{ott2018analyzing}
Myle Ott, Michael Auli, David Grangier, and Marc’Aurelio Ranzato. 2018.
\newblock Analyzing uncertainty in neural machine translation.
\newblock In \emph{International Conference on Machine Learning}, pages 3956--3965. PMLR.

\bibitem[{Ott et~al.(2019)Ott, Edunov, Baevski, Fan, Gross, Ng, Grangier, and Auli}]{ott2019fairseq}
Myle Ott, Sergey Edunov, Alexei Baevski, Angela Fan, Sam Gross, Nathan Ng, David Grangier, and Michael Auli. 2019.
\newblock fairseq: A fast, extensible toolkit for sequence modeling.
\newblock In \emph{Proceedings of NAACL-HLT 2019: Demonstrations}.

\bibitem[{Papineni et~al.(2002)Papineni, Roukos, Ward, and Zhu}]{papineni-etal-2002-bleu}
Kishore Papineni, Salim Roukos, Todd Ward, and Wei-Jing Zhu. 2002.
\newblock \href {https://doi.org/10.3115/1073083.1073135} {{B}leu: a method for automatic evaluation of machine translation}.
\newblock In \emph{Proceedings of the 40th Annual Meeting of the Association for Computational Linguistics}, pages 311--318, Philadelphia, Pennsylvania, USA. Association for Computational Linguistics.

\bibitem[{Peter et~al.(2023)Peter, Vilar, Deutsch, Finkelstein, Juraska, and Freitag}]{peter-etal-2023-theres}
Jan-Thorsten Peter, David Vilar, Daniel Deutsch, Mara Finkelstein, Juraj Juraska, and Markus Freitag. 2023.
\newblock \href {https://doi.org/10.18653/v1/2023.wmt-1.50} {There`s no data like better data: Using {QE} metrics for {MT} data filtering}.
\newblock In \emph{Proceedings of the Eighth Conference on Machine Translation}, pages 561--577, Singapore. Association for Computational Linguistics.

\bibitem[{Popovi{\'c}(2015)}]{popovic-2015-chrf}
Maja Popovi{\'c}. 2015.
\newblock \href {https://doi.org/10.18653/v1/W15-3049} {chr{F}: character n-gram {F}-score for automatic {MT} evaluation}.
\newblock In \emph{Proceedings of the Tenth Workshop on Statistical Machine Translation}, pages 392--395, Lisbon, Portugal. Association for Computational Linguistics.

\bibitem[{Radford et~al.(2023)Radford, Kim, Xu, Brockman, McLeavey, and Sutskever}]{radford2023robust}
Alec Radford, Jong~Wook Kim, Tao Xu, Greg Brockman, Christine McLeavey, and Ilya Sutskever. 2023.
\newblock Robust speech recognition via large-scale weak supervision.
\newblock In \emph{International conference on machine learning}, pages 28492--28518. PMLR.

\bibitem[{Rei et~al.(2022)Rei, Treviso, Guerreiro, Zerva, Farinha, Maroti, C.~de Souza, Glushkova, Alves, Coheur, Lavie, and Martins}]{rei-etal-2022-cometkiwi}
Ricardo Rei, Marcos Treviso, Nuno~M. Guerreiro, Chrysoula Zerva, Ana~C Farinha, Christine Maroti, Jos{\'e}~G. C.~de Souza, Taisiya Glushkova, Duarte Alves, Luisa Coheur, Alon Lavie, and Andr{\'e} F.~T. Martins. 2022.
\newblock \href {https://aclanthology.org/2022.wmt-1.60/} {{C}omet{K}iwi: {IST}-unbabel 2022 submission for the quality estimation shared task}.
\newblock In \emph{Proceedings of the Seventh Conference on Machine Translation (WMT)}, pages 634--645, Abu Dhabi, United Arab Emirates (Hybrid). Association for Computational Linguistics.

\bibitem[{Ren et~al.(2023)Ren, Luo, Zhao, Krishna, Saleh, Lakshminarayanan, and Liu}]{ren2023outofdistribution}
Jie Ren, Jiaming Luo, Yao Zhao, Kundan Krishna, Mohammad Saleh, Balaji Lakshminarayanan, and Peter~J Liu. 2023.
\newblock \href {https://openreview.net/forum?id=kJUS5nD0vPB} {Out-of-distribution detection and selective generation for conditional language models}.
\newblock In \emph{The Eleventh International Conference on Learning Representations}.

\bibitem[{Specia et~al.(2020)Specia, Blain, Fomicheva, Fonseca, Chaudhary, Guzm{\'a}n, and Martins}]{specia-etal-2020-findings-wmt}
Lucia Specia, Fr{\'e}d{\'e}ric Blain, Marina Fomicheva, Erick Fonseca, Vishrav Chaudhary, Francisco Guzm{\'a}n, and Andr{\'e} F.~T. Martins. 2020.
\newblock \href {https://aclanthology.org/2020.wmt-1.79/} {Findings of the {WMT} 2020 shared task on quality estimation}.
\newblock In \emph{Proceedings of the Fifth Conference on Machine Translation}, pages 743--764, Online. Association for Computational Linguistics.

\bibitem[{Team(2024)}]{qwen2.5}
Qwen Team. 2024.
\newblock \href {https://qwenlm.github.io/blog/qwen2.5/} {Qwen2.5: A party of foundation models}.

\bibitem[{Teerapittayanon et~al.(2016)Teerapittayanon, McDanel, and Kung}]{teerapittayanon2016branchynet}
Surat Teerapittayanon, Bradley McDanel, and Hsiang-Tsung Kung. 2016.
\newblock Branchynet: Fast inference via early exiting from deep neural networks.
\newblock In \emph{2016 23rd international conference on pattern recognition (ICPR)}, pages 2464--2469. IEEE.

\bibitem[{Thompson and Post(2020)}]{thompson-post-2020-automatic}
Brian Thompson and Matt Post. 2020.
\newblock \href {https://doi.org/10.18653/v1/2020.emnlp-main.8} {Automatic machine translation evaluation in many languages via zero-shot paraphrasing}.
\newblock In \emph{Proceedings of the 2020 Conference on Empirical Methods in Natural Language Processing (EMNLP)}, pages 90--121, Online. Association for Computational Linguistics.

\bibitem[{Touvron et~al.(2023)Touvron, Lavril, Izacard, Martinet, Lachaux, Lacroix, Rozi{\`e}re, Goyal, Hambro, Azhar et~al.}]{touvron2023llama}
Hugo Touvron, Thibaut Lavril, Gautier Izacard, Xavier Martinet, Marie-Anne Lachaux, Timoth{\'e}e Lacroix, Baptiste Rozi{\`e}re, Naman Goyal, Eric Hambro, Faisal Azhar, et~al. 2023.
\newblock Llama: Open and efficient foundation language models.
\newblock \emph{arXiv preprint arXiv:2302.13971}.

\bibitem[{Tuan et~al.(2021)Tuan, El-Kishky, Renduchintala, Chaudhary, Guzm{\'a}n, and Specia}]{tuan-etal-2021-quality}
Yi-Lin Tuan, Ahmed El-Kishky, Adithya Renduchintala, Vishrav Chaudhary, Francisco Guzm{\'a}n, and Lucia Specia. 2021.
\newblock \href {https://doi.org/10.18653/v1/2021.eacl-main.50} {Quality estimation without human-labeled data}.
\newblock In \emph{Proceedings of the 16th Conference of the European Chapter of the Association for Computational Linguistics: Main Volume}, pages 619--625, Online. Association for Computational Linguistics.

\bibitem[{Wolf(2019)}]{wolf2019huggingface}
T~Wolf. 2019.
\newblock Huggingface's transformers: State-of-the-art natural language processing.
\newblock \emph{arXiv preprint arXiv:1910.03771}.

\bibitem[{Xie et~al.(2024)Xie, Chen, Chen, Peng, Hu, Lin, Peng, Huang, Zhang, Keloth et~al.}]{xie2024me}
Qianqian Xie, Qingyu Chen, Aokun Chen, Cheng Peng, Yan Hu, Fongci Lin, Xueqing Peng, Jimin Huang, Jeffrey Zhang, Vipina Keloth, et~al. 2024.
\newblock Me llama: Foundation large language models for medical applications.
\newblock \emph{arXiv preprint arXiv:2402.12749}.

\bibitem[{Xin et~al.(2020)Xin, Tang, Lee, Yu, and Lin}]{xin-etal-2020-deebert}
Ji~Xin, Raphael Tang, Jaejun Lee, Yaoliang Yu, and Jimmy Lin. 2020.
\newblock \href {https://doi.org/10.18653/v1/2020.acl-main.204} {{D}ee{BERT}: Dynamic early exiting for accelerating {BERT} inference}.
\newblock In \emph{Proceedings of the 58th Annual Meeting of the Association for Computational Linguistics}, pages 2246--2251, Online. Association for Computational Linguistics.

\bibitem[{Yang et~al.(2023)Yang, Meng, Yan, and Zhou}]{yang-etal-2023-rethinking}
Zhen Yang, Fandong Meng, Yuanmeng Yan, and Jie Zhou. 2023.
\newblock \href {https://doi.org/10.18653/v1/2023.findings-acl.126} {Rethinking the word-level quality estimation for machine translation from human judgement}.
\newblock In \emph{Findings of the Association for Computational Linguistics: ACL 2023}, pages 2012--2025, Toronto, Canada. Association for Computational Linguistics.

\bibitem[{Yuan et~al.(2021)Yuan, Neubig, and Liu}]{NEURIPS2021_e4d2b6e6}
Weizhe Yuan, Graham Neubig, and Pengfei Liu. 2021.
\newblock \href {https://proceedings.neurips.cc/paper/2021/file/e4d2b6e6fdeca3e60e0f1a62fee3d9dd-Paper.pdf} {Bartscore: Evaluating generated text as text generation}.
\newblock In \emph{Advances in Neural Information Processing Systems}, volume~34, pages 27263--27277. Curran Associates, Inc.

\bibitem[{Zaragoza-Bernabeu et~al.(2022)Zaragoza-Bernabeu, Ram{\'i}rez-S{\'a}nchez, Ba{\~n}{\'o}n, and Ortiz~Rojas}]{zaragoza-bernabeu-etal-2022-bicleaner}
Jaume Zaragoza-Bernabeu, Gema Ram{\'i}rez-S{\'a}nchez, Marta Ba{\~n}{\'o}n, and Sergio Ortiz~Rojas. 2022.
\newblock \href {https://aclanthology.org/2022.lrec-1.87/} {Bicleaner {AI}: Bicleaner goes neural}.
\newblock In \emph{Proceedings of the Thirteenth Language Resources and Evaluation Conference}, pages 824--831, Marseille, France. European Language Resources Association.

\bibitem[{Zerva et~al.(2024)Zerva, Blain, C.~De~Souza, Kanojia, Deoghare, Guerreiro, Attanasio, Rei, Orasan, Negri, Turchi, Chatterjee, Bhattacharyya, Freitag, and Martins}]{zerva-etal-2024-findings}
Chrysoula Zerva, Frederic Blain, Jos{\'e}~G. C.~De~Souza, Diptesh Kanojia, Sourabh Deoghare, Nuno~M. Guerreiro, Giuseppe Attanasio, Ricardo Rei, Constantin Orasan, Matteo Negri, Marco Turchi, Rajen Chatterjee, Pushpak Bhattacharyya, Markus Freitag, and Andr{\'e} Martins. 2024.
\newblock \href {https://doi.org/10.18653/v1/2024.wmt-1.3} {Findings of the quality estimation shared task at {WMT} 2024: Are {LLM}s closing the gap in {QE}?}
\newblock In \emph{Proceedings of the Ninth Conference on Machine Translation}, pages 82--109, Miami, Florida, USA. Association for Computational Linguistics.

\bibitem[{Zhang et~al.(2019)Zhang, Kishore, Wu, Weinberger, and Artzi}]{zhang2019bertscore}
Tianyi Zhang, Varsha Kishore, Felix Wu, Kilian~Q Weinberger, and Yoav Artzi. 2019.
\newblock Bertscore: Evaluating text generation with bert.
\newblock \emph{arXiv preprint arXiv:1904.09675}.

\bibitem[{Zheng et~al.(2023)Zheng, Chiang, Sheng, Zhuang, Wu, Zhuang, Lin, Li, Li, Xing et~al.}]{zheng2023judging}
Lianmin Zheng, Wei-Lin Chiang, Ying Sheng, Siyuan Zhuang, Zhanghao Wu, Yonghao Zhuang, Zi~Lin, Zhuohan Li, Dacheng Li, Eric Xing, et~al. 2023.
\newblock Judging llm-as-a-judge with mt-bench and chatbot arena.
\newblock \emph{Advances in neural information processing systems}, 36:46595--46623.

\bibitem[{Zouhar et~al.(2023)Zouhar, Dhuliawala, Zhou, Daheim, Kocmi, Jiang, and Sachan}]{zouhar-etal-2023-poor}
Vil{\'e}m Zouhar, Shehzaad Dhuliawala, Wangchunshu Zhou, Nico Daheim, Tom Kocmi, Yuchen~Eleanor Jiang, and Mrinmaya Sachan. 2023.
\newblock \href {https://doi.org/10.18653/v1/2023.eacl-main.95} {Poor man`s quality estimation: Predicting reference-based {MT} metrics without the reference}.
\newblock In \emph{Proceedings of the 17th Conference of the European Chapter of the Association for Computational Linguistics}, pages 1311--1325, Dubrovnik, Croatia. Association for Computational Linguistics.

\bibitem[{Zouhar et~al.(2024)Zouhar, Ding, Currey, Badeka, Wang, and Thompson}]{zouhar-etal-2024-fine}
Vil{\'e}m Zouhar, Shuoyang Ding, Anna Currey, Tatyana Badeka, Jenyuan Wang, and Brian Thompson. 2024.
\newblock \href {https://doi.org/10.18653/v1/2024.acl-short.45} {Fine-tuned machine translation metrics struggle in unseen domains}.
\newblock In \emph{Proceedings of the 62nd Annual Meeting of the Association for Computational Linguistics (Volume 2: Short Papers)}, pages 488--500, Bangkok, Thailand. Association for Computational Linguistics.

\end{thebibliography}
% Custom bibliography entries only
% \bibliography{custom}

\newpage

\appendix

\section{Human Analysis of Valid Output Tokens} \label{sec:human}
\begin{figure}[htbp]
    \centering
    \includegraphics[width=\linewidth]{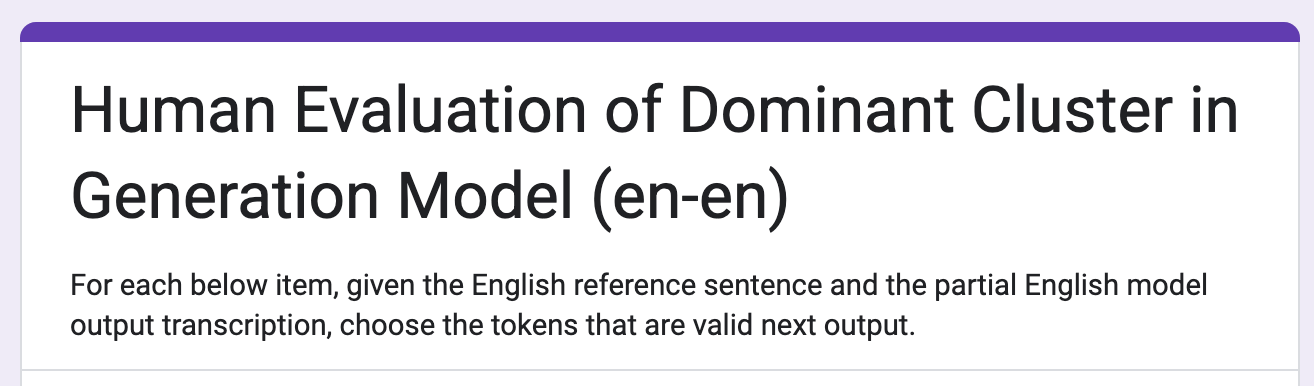} \vspace{0.1cm}

    \includegraphics[width=\linewidth]{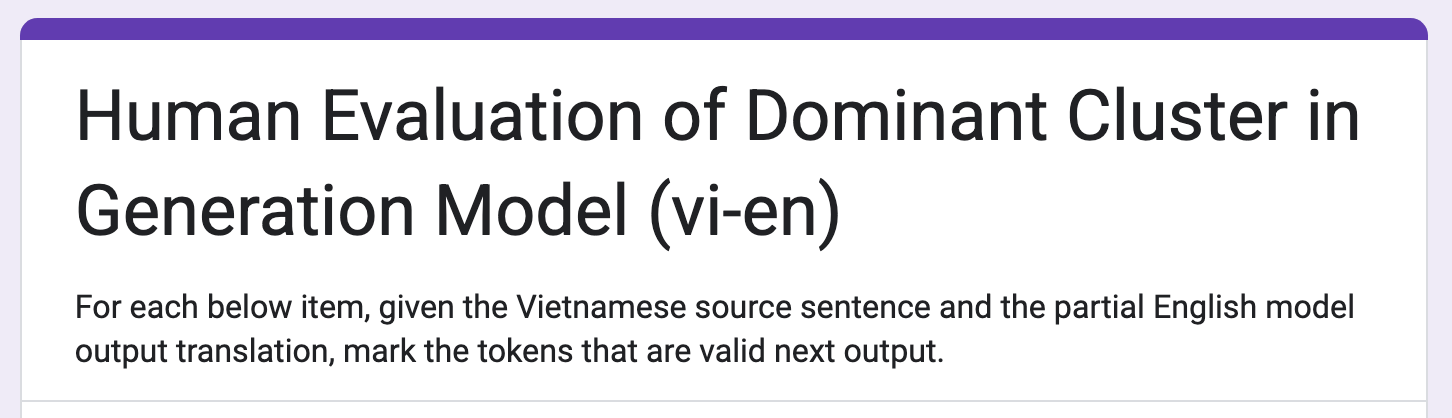} \vspace{0.1cm}

    \includegraphics[width=\linewidth]{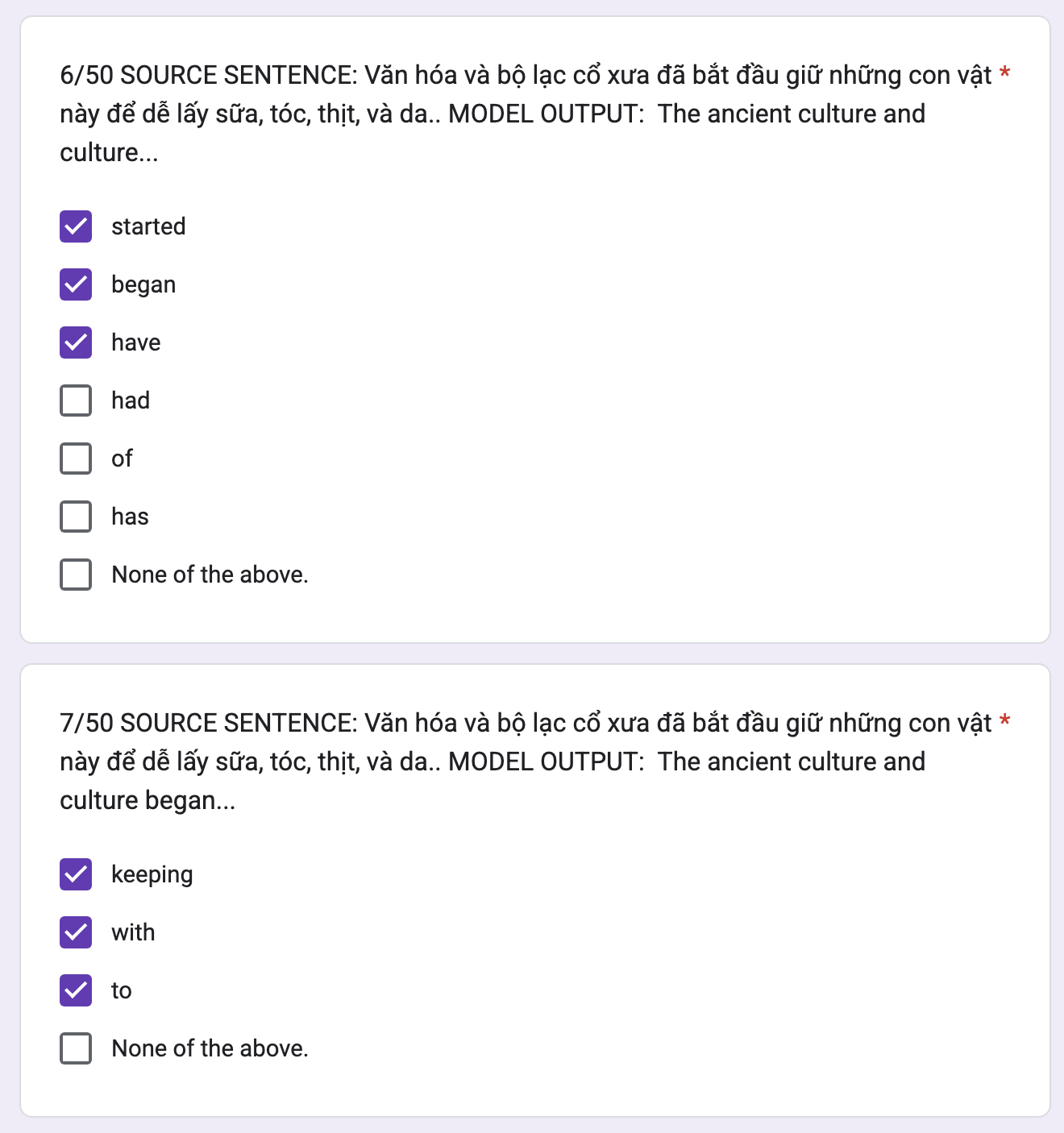}
    \caption{Snapshots of the forms provided to human annotators.}
    \label{fig:humanForm}
\end{figure}

We perform a human analysis on the number of valid tokens at each output step. We consider the ambiguous Speech Translation (ST) task on Vietnamese-English, and the less-ambiguous Automatic Speech Recognition (ASR) task on English. We use Whisper Larger V3 output on the Fleurs test set. We identify the dominant tokens at each output step using our approach presented in Section \ref{sec:method}, show them to the annotators, and ask the annotators to mark which tokens are the correct output. Figure \ref{fig:humanForm} contains snapshots of the forms we gave to the annotators.

In this study, we presented 50 samples to two annotators and reported their average responses. Both annotators are students, native Vietnamese speakers with undergraduate and postgraduate education conducted in English, ensuring high proficiency in both languages. They voluntarily participated in this study without payment, and agreed to their responses being published in this paper.

The results on the number of valid tokens at each output step are already discussed in Section \ref{sec:motivation}, Figure \ref{fig:asr_st_nr_correct}. Additionally, we calculate the portion of tokens within the dominant clusters that are actually valid output. We obtained 89.02\% for the ST task, and 66.86\% for the ASR task. This again shows the effectiveness of our approach to identify tokens that are important within the output distributions for ambiguous tasks, and its weaker usage for less ambiguous tasks.

\section{Theoretical Analysis} \label{app:theoretical}
Language models generate text by predicting the next token $y_t$ given the previous context $y_{<t}$ and the input $X$, using conditional probability:

\[
P(y_t | y_{<t}, X) = softmax(z_t)
\]

where $ z_t \in R^{|V|} $ is the logit vector at output step $t$ over the vocabulary $V$.

The softmax function defines the probability of the selected token $w_i \in V$ as:

\[
P(y_t = w_i | y_{<t}, X) = \frac{e^{z_{t,i}}}{\sum_{j=1}^{|V|} e^{z_{t,j}}}
\]

The softmax function satisfies:
$P(y_t = w_i) \geq 0$ and $\sum_{i=1}^{|V|} P(y_t = w_i) = 1$.

In natural language, multiple tokens can be the valid next output. Let:
\begin{itemize}
    \item $C = \{w_{c1}, w_{c2}, ...\} \subset V$: the set of correct tokens at step $t$
    \item $|C| = k$: number of correct tokens.
\end{itemize}

If we want each correct token to have a minimum probability $p_{min}$, then:

\[
\sum_{i \in C} P(y_t = w_i) \geq k \cdot p_{min}
\]

However, since $ \sum_{i \in V} P(y_t = w_i) = 1 $, we must have:

\[
k \cdot p_{min} \leq 1 \quad \Rightarrow \quad p_{min} \leq \frac{1}{k}
\]

Thus, as $k$ increases, the maximum assignable probability to each correct token decreases, leading to underconfidence.

With \textsc{BoostedProb}, we instead use the sum of the probability mass of $C$ as the quality score:

\begin{align*}
    & BooostedProb(y_t = w_{c1}) \\
    &= BooostedProb(y_t = w_{c2}) \\
    &= ... \\
    &= P_C = \sum_{i \in C} P(y_t = w_i | y_{<t}, X)
\end{align*}

The only condition on $P_C$ is that $P_C \leq 1$. Therefore, with BoostedProb, $p_{min}$ can have a high value close to one, regardless of the size of $C$, thus tackling the underconfidence issue.

\section{Hyperparameters Tuning} \label{sec:hyperparamtune}
\subsection{Finding Dominant Tokens}
We tune the hyperparameters for our approach, i.e., the value of $x$\% that defines the relative threshold, and the value of $\epsilon$ that defines the absolute threshold that makes a reduction in probability mass significant, thus separating dominant from non-dominant tokens. The set of candidate values for $x$\% is $\{0.2, 0.3, 0.4, 0.5, 0.6\}$. The set of candidate values for $\epsilon$ is $\{0.005, 0.01, 0.1\}$.

\begin{figure}[h]
    \centering
    \includegraphics[width=\linewidth]{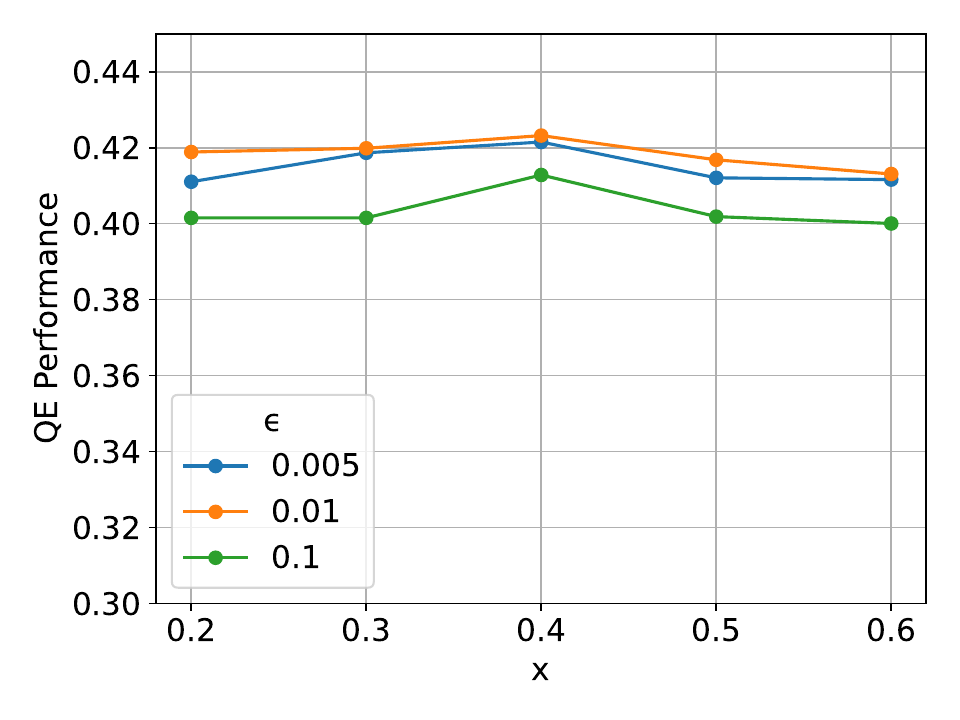}
    \caption{Hyperparameter tuning on Whisper Large V3.}
    \label{fig:htuneWhisper}
\end{figure}

\begin{figure}[h]
    \centering
    \includegraphics[width=\linewidth]{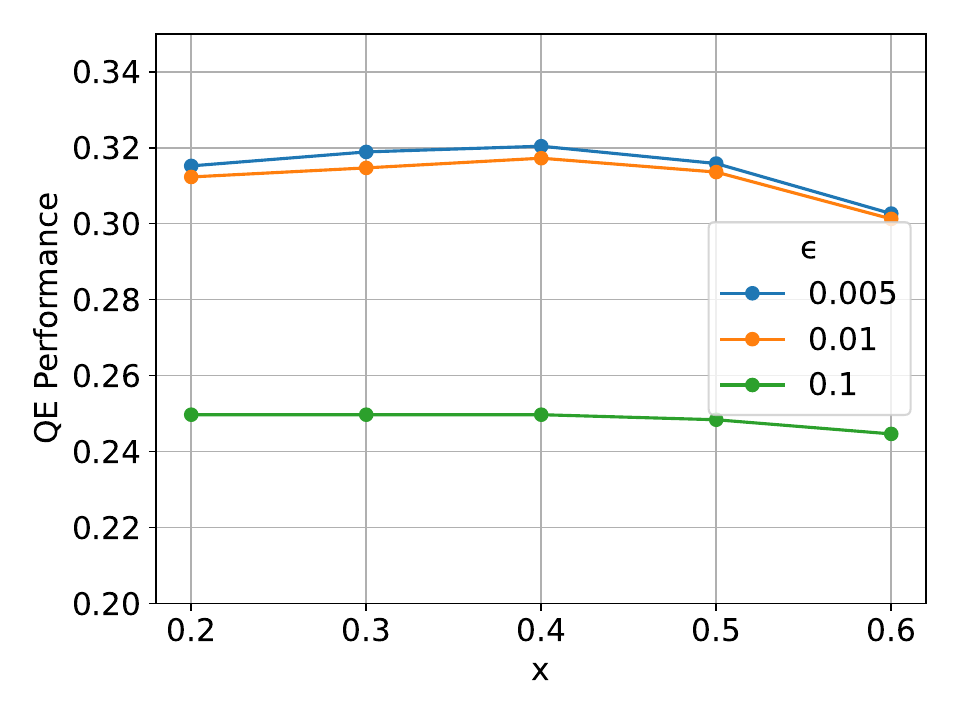}
    \caption{Hyperparameter tuning on DeltaLM Large.}
    \label{fig:htuneDeltaLM}
\end{figure}

\begin{figure}[h]
    \centering
    \includegraphics[width=\linewidth]{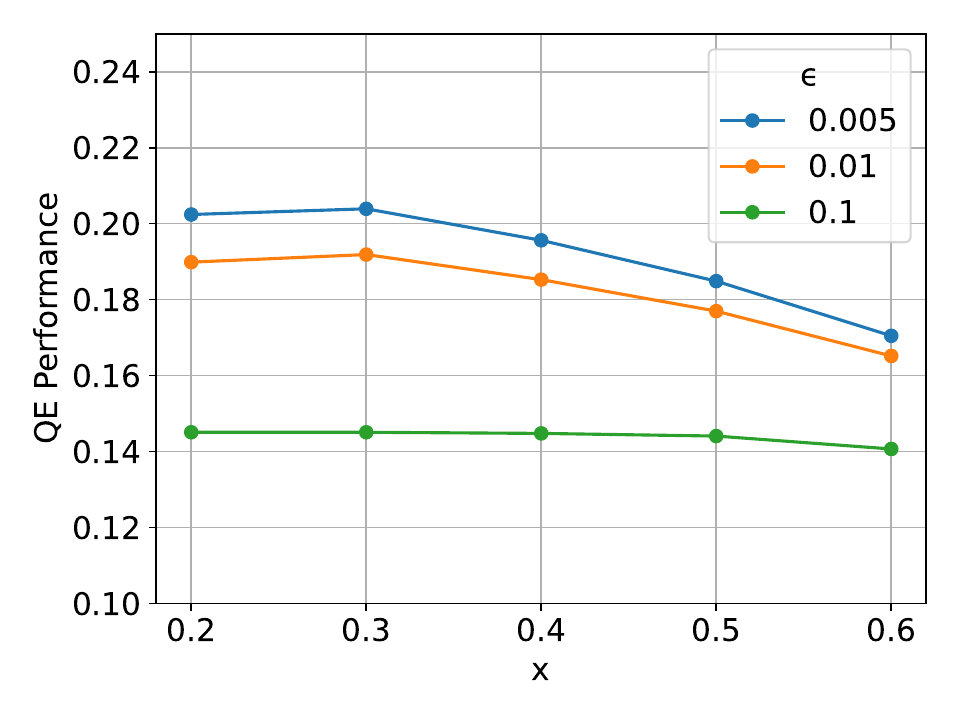}
    \caption{Hyperparameter tuning on Tower.}
    \label{fig:htuneTower}
\end{figure}

We perform hyperparameter tuning on the development splits of the datasets: 5,000 samples from ParaCrawl for each of the language pairs \textit{en-de} and \textit{zh-en} for Machine Translation, and the pre-defined development split of Fleurs on 4 language pairs \textit{de-en, es-en, vi-en, zh-en} for Speech Translation. We tune for three models: Whisper Large V3, DeltaLM and Tower, take the average over all language pairs, and report the results in Figure \ref{fig:htuneWhisper}, Figure \ref{fig:htuneDeltaLM} and Figure \ref{fig:htuneTower}, respectively.

\FloatBarrier

As can be seen from the plots, for all models, the hyperparameter set with $x=0.3$ and $\epsilon=0.005$ gives the best or close to the best performance. This shows that our approach is rather robust to hyperparameters. As a result, we use these values for all of our experiments in the paper.

\subsection{Sequence Aggregation of Token Scores}
We also experiment with different ways to aggregate token-level scores to sequence-level scores with our approach. We try taking the mean, the median, and the minimum of the token-level scores. We also try counting the number of output tokens in the sequence that are within the dominant clusters, rather than using the probability mass of the clusters. 

The results are shown in Table \ref{tab:aggregation}. As can be seen, using the probability mass is better than simply counting the number of dominant tokens. Taking the mean of the token scores is better than taking the median, potentially due to the median ignoring catastrophic token-level errors. Taking the minimum of the token scores ended up giving better performance than taking the mean, indicating that the quality of the lowest-score token might be sufficient to represent the quality of the whole sequence. However, we stick with using the mean as the token-to-sequence score aggregation method in our experiments, in order to have a more global view of the whole sequence quality.

\begin{table}[htbp]
\small
\centering
\begin{tabular}{lcccc}
             & Whisper & DeltaLM & Tower & \textbf{Avg.}  \\ \hline
Mean         & 0.419   & 0.319   & 0.204 & 0.314          \\
Median       & 0.272   & 0.253   & 0.147 & 0.224          \\
Min          & 0.439   & 0.436   & 0.236 & \textbf{0.370} \\
Nr. Dominant & 0.107   & 0.315   & 0.216 & 0.213         
\end{tabular}
\caption{Different token-to-sequence scrore aggregation methods.}
\label{tab:aggregation}
\end{table}

\section{Randomness in Inference} \label{app:std}
To account for randomness in the inference process, it is ideal to repeat each experiment with different random seeds and report the variance. However, this approach is computationally expensive. In our work, we instead prioritize running a broad set of experiments across models and tasks, rather than repeating each experiment multiple times. For illustration, we conduct an additional study with multiple seeds in one setting: the NLLB model on a translation task. Specifically, we apply top-$p$ sampling with $p=0.5$ and repeat the experiment using five different seeds (integers from 0 to 4). The results, which incorporate performance variance, are presented in Table~\ref{tab:std} and demonstrate that our approach consistently outperforms the baselines.

\begin{table}[htbp]
\small
\centering
\setlength{\tabcolsep}{4pt}
\begin{tabular}{lccc}
 Lang. & Probability          & Entropy           & \textsc{BoostedProb}  \\ \hline
 en-de & 0.142 ± {\color{gray}0.002} & 0.449 ± {\color{gray}0.030} & \textbf{0.481} ± {\color{gray}0.039} \\
 zh-en & 0.184 ± {\color{gray}0.004} & 0.226 ± {\color{gray}0.010} & \textbf{0.312} ± {\color{gray}0.010}                
\end{tabular}
\caption{Performance of QE methods in Pearson correlation across runs with different random seed on NLLB translation.}
\label{tab:std}
\end{table}

\section{Correlation with ROUGE-L and BertScore} \label{sec:more_ref}

\begin{table*}[htbp]
\small
\centering
\setlength{\tabcolsep}{4pt}
\begin{tabular}{llllcccc}
Ground Truth   & Task               & Model        & Test Set & Language & Probability    & Entropy        & BoostedProb    \\ \hline
Bart Score     & Summarization      & Bloomz 560M  & XSum     & en       & -0.003         & 0.176          & \textbf{0.210} \\
               &                    & Llama3.2 3B  & XSum     & en       & 0.002          & 0.201          & \textbf{0.209} \\
               &                    & Llama3.3 70B & XSum     & en       & 0.001          & 0.000          & \textbf{0.004} \\
               & Question Answering & Bloomz 560M  & GSM8K    & en       & -0.002         & \textbf{0.111} & 0.009          \\
               & (Math)             & Llama3.2 3B  & GSM8K    & en       & -0.007         & 0.006          & \textbf{0.111} \\
               &                    & Llama3.3 70B & GSM8K    & en       & -0.001         & 0.005          & \textbf{0.006} \\
               & Question Answering & Bloomz 560M  & SciEx    & en,de    & -0.002         & 0.005          & \textbf{0.006} \\
               & (University Exam)  & Llama3.2 3B  & SciEx    & en,de    & 0.002          & 0.228          & \textbf{0.310} \\
               &                    & Llama3.3 70B & SciEx    & en,de    & 0.103          & \textbf{0.180} & \textbf{0.180} \\
               &                    &              &          &          &                &                &                \\
RougeL         & Summarization      & Bloomz 560M  & XSum     & en       & 0.048          & 0.176          & \textbf{0.216} \\
               &                    & Llama3.2 3B  & XSum     & en       & -0.207         & \textbf{0.632} & 0.619          \\
               &                    & Llama3.3 70B & XSum     & en       & 0.061          & 0.001          & \textbf{0.061} \\
               & Question Answering & Bloomz 560M  & GSM8K    & en       & 0.049          & 0.107          & \textbf{0.125} \\
               & (Math)             & Llama3.2 3B  & GSM8K    & en       & -0.169         & -0.054         & \textbf{0.079} \\
               &                    & Llama3.3 70B & GSM8K    & en       & 0.148          & \textbf{0.294} & 0.227          \\
               & Question Answering & Bloomz 560M  & SciEx    & en,de    & -0.273         & 0.473          & \textbf{0.497} \\
               & (University Exam)  & Llama3.2 3B  & SciEx    & en,de    & 0.102          & 0.138          & \textbf{0.141} \\
               &                    & Llama3.3 70B & SciEx    & en,de    & 0.182          & 0.204          & \textbf{0.223} \\
               &                    &              &          &          &                &                &                \\
Bert Score     & Summarization      & Bloomz 560M  & XSum     & en       & -0.083         & 0.023          & \textbf{0.111} \\
               &                    & Llama3.2 3B  & XSum     & en       & -0.121         & 0.091          & \textbf{0.099} \\
               &                    & Llama3.3 70B & XSum     & en       & 0.070          & -0.024         & \textbf{0.162} \\
               & Question Answering & Bloomz 560M  & GSM8K    & en       & -0.022         & -0.022         & -0.038         \\
               & (Math)             & Llama3.2 3B  & GSM8K    & en       & -0.121         & -0.257         & -0.126         \\
               &                    & Llama3.3 70B & GSM8K    & en       & \textbf{0.121} & -0.361         & -0.257         \\
               & Question Answering & Bloomz 560M  & SciEx    & en,de    & 0.420          & 0.411          & \textbf{0.434} \\
               & (University Exam)  & Llama3.2 3B  & SciEx    & en,de    & 0.009          & 0.007          & \textbf{0.015} \\
               &                    & Llama3.3 70B & SciEx    & en,de    & 0.005          & 0.164          & \textbf{0.380} \\
               &                    &              &          &          &                &                &                \\
LLM-as-a-Judge & Summarization      & Bloomz 560M  & XSum     & en       & 0.014          & 0.265          & \textbf{0.287} \\
               &                    & Llama3.2 3B  & XSum     & en       & -0.039         & 0.000          & \textbf{0.156} \\
               &                    & Llama3.3 70B & XSum     & en       & 0.018          & 0.130          & \textbf{0.184} \\
               & Question Answering & Bloomz 560M  & GSM8K    & en       & 0.002          & 0.023          & -0.013         \\
               & (Math)             & Llama3.2 3B  & GSM8K    & en       & 0.015          & -0.204         & \textbf{0.128} \\
               &                    & Llama3.3 70B & GSM8K    & en       & 0.024          & -0.035         & \textbf{0.053} \\
               & Question Answering & Bloomz 560M  & SciEx    & en,de    & 0.539          & 0.646          & \textbf{0.654} \\
               & (University Exam)  & Llama3.2 3B  & SciEx    & en,de    & -0.030         & -0.123         & -0.008         \\
               &                    & Llama3.3 70B & SciEx    & en,de    & 0.076          & 0.229          & \textbf{0.305} \\
               &                    &              &          &          &                &                &               
\end{tabular}
\caption{Correlation of different quality scores to different ground truth metrics: Bart Score, RougeL, BertScore, and LLM-as-a-Judge with Qwen2.5 72B.}
\label{tab:more_ref}
\end{table*}

Since relying on a single reference-based metric - specifically BARTScore in our main experiments - may not provide a sufficiently robust ground truth for evaluating reference-free Quality Estimation approaches, we additionally report results using ROUGE-L \cite{lin-och-2004-automatic} and BERTScore \cite{zhang2019bertscore}. We also repeat the results on Bart Score, which we presented in Section \ref{sec:results_overall}, for a more complete overview. 

The results are shown in Table \ref{tab:more_ref}. With all three ground truth metrics, we generally observe similar patterns: our \textsc{BoostedProb} approach performs better than the raw probability and the probability entropy. One exception is on the Question Answering task evaluated with Bert Score as ground truth, where all QE approaches give a negative correlation most of the time. This is possible due to the GSM8k test set is about solving math problems. Bert Score might not be a suitable metric, since it compares the contextual embeddings of the model output to the reference, which might not emphasize critical errors with wrong number output in math problems, since numbers might be close to each other in the embedding space.

\section{Negative Effect on Very Weak Models} \label{sec:negative}
As mentioned in Section \ref{sec:fail}, \textsc{BoostedProb} improves QE performance of output probability for stronger models, but worsens it for weak models. This is somewhat expected, since the motivation of \textsc{BoostedProb} is to improve cases when the model is underconfident. It does not consider the cases when a low-quality model is overconfident and constantly assigns high probability values to the wrong token. To test whether this is truly the cause, we manually look at some output by the worst-performing model, Whisper Tiny, on Chinese-to-English test data. One example is as follows:\\
\textbf{\textit{Source}}:
\begin{CJK*}{UTF8}{gbsn}
\small 
 "\textit{有了它，我们才有了火车、汽车和许多其他交通工具}"
\end{CJK*}\\
\textbf{\textit{Reference}}: "\textit{It has brought us the train, the car, and many other transportation devices.}"\\
\textbf{\textit{Model output}}: "\textit{There we have it.}"

\begin{figure}[htbp]
    \centering
    \includegraphics[width=0.7\linewidth]{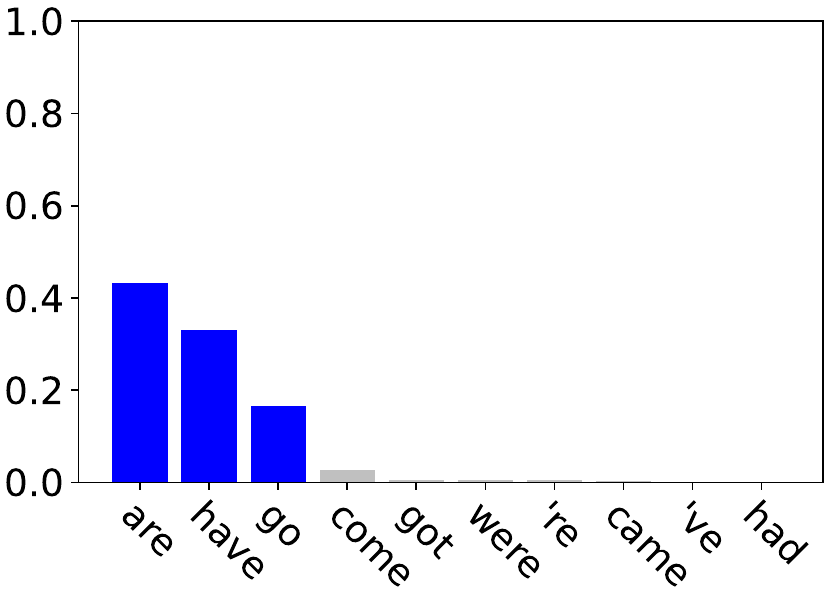}
    \caption{Example of Whisper Tiny's hallucinated probability distribution at an output step.}
    \label{fig:example_hallu}
\end{figure}

Observe that the model exhibits signs of hallucination, as the output is quite irrelevant to the input sentence and the ground-truth reference. However, when we look at the probability distributions of the output tokens, they do form dominant clusters. For example, at the third output step after "\textit{There we ...}", the dominant next tokens assigned by the model are "\textit{are}", "\textit{have}" and "\textit{go}", as shown in Figure \ref{fig:example_hallu}. These tokens seem to be hallucinated: they are common words that might come after  "\textit{There we ...}", but are quite irrelevant to the input sentence. In cases like this, by favoring the dominant tokens, our approach emphasizes the models' overconfidence, thus leading to bad quality estimation performance.

\section{Effect on Less Ambiguous Tasks} \label{sec:asr}
We expect that \textsc{BoostedProb} is unlikely to have major differences from raw model probability on less ambiguous tasks, such as Automatic Speech Recognition (ASR). Recall that for ASR, the size of the dominant clusters is usually 1 (see Section \ref{sec:motivation}, Figure \ref{fig:asr_st_nr_dominant}). Therefore, using the total probability mass of the dominant cluster in \textsc{BoostedProb} would be the same as using the raw model probability of the single dominant token.

We confirm this hypothesis by applying \textsc{BoostedProb} on the ASR task with Whisper Large V3 and the Fleurs test set, as shown in Table \ref{tab:asr}. The QE performance here is the Pearson correlation of the QE scores with the Word Error Rate of the transcription output. As can be seen, the QE performance of \textsc{BoostedProb} is similar to that of the raw model probability.

\begin{table}[htbp]
\centering
\small
\begin{tabular}{lcccc}
            & en    & de    & es    & zh    \\ \hline
Probability & 0.363 & 0.346 & 0.375 & 0.375 \\
\textsc{BoostedProb} & 0.364 & 0.378 & 0.383 & 0.396
\end{tabular}
\caption{\textsc{BoostedProb} versus raw probability on ASR tasks.}
\label{tab:asr}
\end{table}

We can conclude that \textsc{BoostedProb} gives the same or better QE performance than the raw model probability, depending on the magnitude of ambiguity of the task at hand.

\section{Discussion on Overall QE correlations}
As can be seen in Section \ref{sec:results_overall} and Section \ref{sec:prism_results}, our method gives around 0.3 points on average, and 0.688 at max in Pearson correlation with the gold quality score. One can raise the question: Can this be interpreted as correlated at all? Does this mean that the proposed approach (as well as the baselines) offers very limited practical use?

\begin{figure}[htbp]
    \centering
    \includegraphics[width=\linewidth]{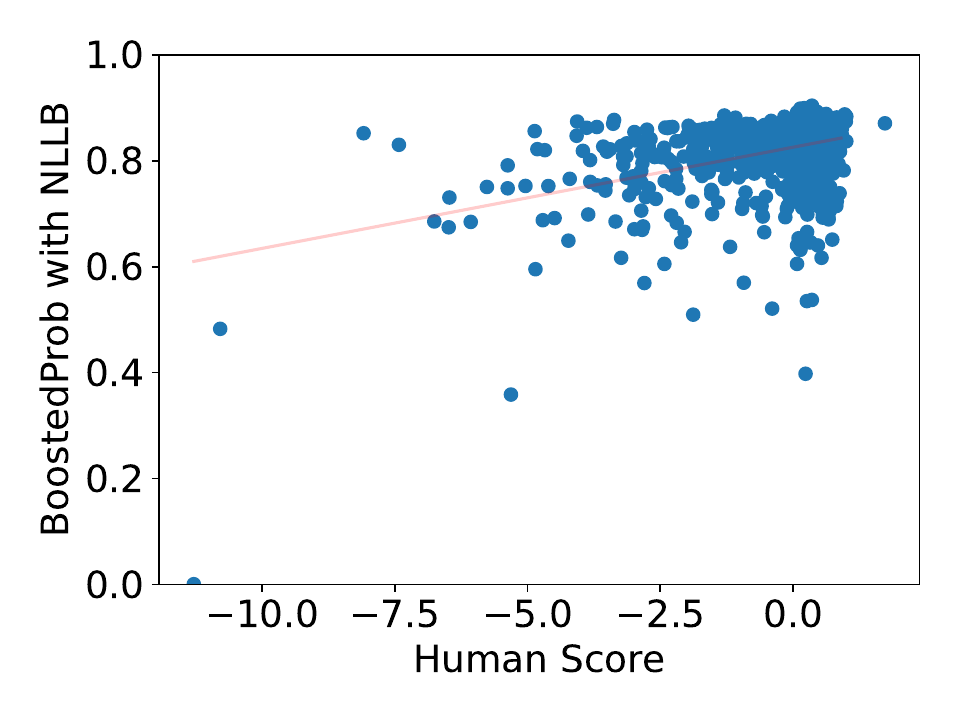}
    \caption{Illustration of the positive correlation between our method on top of Prism with NLLB and the human score, on the WMT22 General data, \textit{en-de}.}
    \label{fig:corr0384}
\end{figure}

Note that Pearson correlation ranges between -1 and 1. An example plot of our BoostedProb approach with NLLB on Prism, in correlation with human scores, on the WMT 22 \textit{en-de} data is shown in Figure \ref{fig:corr0384}. With Pearson correlation of 0.384, we can already see the positive trend from the plot.

As can be seen in Section \ref{sec:prism_results}, even the supervised Quality Estimation model obtained around 0.4 correlation. As a broader pointer, we can consider the results of a recent public shared task on Quality Estimation for Machine Translation at WMT 2024 \cite{zerva-etal-2024-findings}. In their findings, page 103, appendix B, Table 8, the Pearson correlations of the participating QE systems are also around 0.4, including the SOTA system by Unbabel. 

Even with this current progress of the field, Quality Estimation has already shown to be useful in many applications, e.g., guiding the decoding process to generate better translation \cite{fernandes-etal-2022-quality}, supporting post-editing \cite{deoghare-etal-2023-quality}, or to filter out synthetically created bilingual data \cite{peter-etal-2023-theres}.

\section{Tools and Hardwares}
The Speech Translation experiments are conducted using Huggingface \cite{wolf2019huggingface}. The Text Translation experiments are conducted using Fairseq \cite{ott2019fairseq}. The Summarization and Question Answering experiments are conducted using LM-Polygraph \cite{fadeeva-etal-2023-lm}. For all experiments, we use A100 GPUs with 40GB of memory.

\section{License For Artifacts}
The license for artifacts used in our paper is as follows:

\begin{itemize}
    \item Fleurs dataset \cite{conneau2023fleurs}: CC BY 4.0
    \item ParaCrawl dataset \cite{banon-etal-2020-paracrawl}: Creative Commons CC0
    \item WMT22 General dataset \cite{kocmi-etal-2022-findings}: Apache License 2.0
    \item XSum dataset \cite{narayan-etal-2018-dont}: MIT License
    \item GSM8k dataset \cite{cobbe2021training}: MIT License
    \item Whisper models \cite{radford2023robust}: Apache License 2.0
    \item DeltaLM model \cite{ma2021deltalm}: MIT License
    \item NLLB model \cite{costa2022no}: CC BY NC 4.0
    \item Tower model \cite{tower_llm_2024}: CC BY NC 4.0
    \item Bloomz model \cite{muennighoff-etal-2023-crosslingual}: The BigScience RAIL License
    \item Llama 3.2 models \cite{touvron2023llama}: Llama 3.2 Community License Agreement
    \item Llama 3.3 models \cite{touvron2023llama}: Llama 3.3 Community License Agreement
\end{itemize}

\end{document}